\documentclass{article}

\usepackage{microtype}
\usepackage{graphicx}
\usepackage{subfigure}
\usepackage{booktabs} 

\usepackage{hyperref}
\usepackage{url}
\usepackage{graphicx}
\usepackage{soul}
\usepackage{xcolor}
\usepackage{amssymb}
\usepackage{bbm}
\usepackage{amsmath}
\usepackage{mathtools}
\usepackage{tabularx}
\usepackage{subfiles}
\usepackage{bm}
\usepackage{natbib}
\usepackage{algorithm}
\usepackage{algorithmic}

\usepackage[accepted]{icml2024}

\usepackage{amsmath}
\usepackage{amssymb}
\usepackage{mathtools}
\usepackage{amsthm}

\usepackage[capitalize,noabbrev]{cleveref}

\usepackage{CJKutf8}

\theoremstyle{plain}

\theoremstyle{definition}

\theoremstyle{remark}

\usepackage[textsize=tiny]{todonotes}

\newcommand{\bigcell}[2]{\begin{tabular}[t]{@{}#1@{}}#2\end{tabular}}

\newcommand{\btheta}{{\bm \theta}}
\newcommand{\bx}{{\bf x}}
\newcommand{\by}{{\bf y}}

\icmltitlerunning{Simulation-Based Inference with Quantile Regression}

\begin{document}

\twocolumn[
\icmltitle{Simulation-Based Inference with Quantile Regression}




\begin{icmlauthorlist}
\icmlauthor{He Jia$\ $(\begin{CJK*}{UTF8}{gbsn}贾赫\end{CJK*})}{yyy}
\end{icmlauthorlist}

\icmlaffiliation{yyy}{Department of Astrophysical Sciences, Princeton University, USA}

\icmlcorrespondingauthor{He Jia}{hejia@princeton.edu}

\icmlkeywords{Machine Learning, ICML}

\vskip 0.3in
]



\printAffiliationsAndNotice{}  

\begin{abstract}
We present Neural Quantile Estimation (NQE), a novel Simulation-Based Inference (SBI) method based on conditional quantile regression.
NQE autoregressively learns individual one dimensional quantiles for each posterior dimension, conditioned on the data and previous posterior dimensions.
Posterior samples are obtained by interpolating the predicted quantiles using monotonic cubic Hermite spline, with specific treatment for the tail behavior and multi-modal distributions.
We introduce an alternative definition for the Bayesian credible region using the local Cumulative Density Function (CDF), offering substantially faster evaluation than the traditional Highest Posterior Density Region (HPDR).
In case of limited simulation budget and/or known model misspecification, a post-processing calibration step can be integrated into NQE to ensure the unbiasedness of the posterior estimation with negligible additional computational cost.
We demonstrate that NQE achieves state-of-the-art performance on a variety of benchmark problems.
\end{abstract}

\section{Introduction}

\label{sec:intro}


Given the likelihood $p(\bx|\btheta)$ of a stochastic forward model and observation data $\bf{x}$, Bayes' theorem postulates that the underlying model parameters $\btheta$ follow the posterior distribution $p(\btheta|\bx)\propto p(\bx|\btheta)p(\btheta)$, where $p(\btheta)$ represents the prior.
In many applications, however, we are restricted to simulating the data $\bx\sim p(\bx|\btheta)$, while the precise closed form of $p(\bx|\btheta)$ remains unavailable.
Simulation-Based Inference (SBI), also known as Likelihood-Free Inference (LFI) or Implicit Likelihood Inference (ILI), conducts Bayesian inference directly from these simulations, circumventing the need to explicitly formulate a tractable likelihood function.
Early research in this field primarily consists of Approximate Bayesian Computation (ABC) variants, which employ a distance metric in the data space and approximate true posterior samples using realizations whose simulated data are ``close enough" to the observation \citep[e.g.][]{tavare1997inferring,pritchard1999population,beaumont2002approximate,beaumont2009adaptive}. However, these methods are prone to the curse of dimensionality and prove inadequate for higher-dimensional applications.

In recent years, a series of neural-network-based SBI methods have been proposed, which can be broadly categorized into three groups.
Neural Likelihood Estimation \citep[NLE,][]{papamakarios2019sequential,lueckmann2019likelihood} fits the likelihood using a neural density estimator, typically based on Normalizing Flows.
The posterior is then evaluated by multiplying the likelihood with the prior, and posterior samples can be drawn using Markov Chain Monte Carlo (MCMC).
Neural Posterior Estimation \citep[NPE,][]{papamakarios2016fast,lueckmann2017flexible,greenberg2019automatic} uses neural density estimators to approximate the posterior, thereby enabling direct posterior sample draws without running MCMC.
Neural Ratio Estimation \citep[NRE,][]{hermans2020likelihood} employs classifiers to estimate density ratios, commonly selected as the likelihood-to-evidence ratio.
Indeed, \citet{durkan2020contrastive} demonstrates that NRE can be unified with specific types of NPE under a general contrastive learning framework.
Each method has its sequential counterpart, namely SNLE, SNPE, and SNRE, respectively.
Whereas standard NLE, NPE, and NRE allocate new simulations based on the prior, allowing them to be applied to any observation data (i.e., they are \textit{amortized}), their sequential counterparts allocate new simulations based on the inference results from previous iterations and must be trained specifically for each observation.
These neural-network-based methods typically surpass traditional ABC methods in terms of inference accuracy under given simulation budgets.
See \citet{cranmer2020frontier} for a review and \citet{lueckmann2021benchmarking} for a comprehensive benchmark of prevalent SBI methods.

Quantile Regression (QR), as introduced by \citet{koenker1978regression}, estimates the conditional quantiles of the response variable over varying predictor variables.
Many Machine Learning (ML) algorithms can be extended to quantile regression by simply transitioning to a weighted $L_1$ loss \citep[e.g.][]{meinshausen2006quantile,rodrigues2020beyond,tang2022nonparametric}.
In this paper, we introduce Neural Quantile Estimation (NQE), a new family of SBI methods supplementing the existing NPE, NRE and NLE approaches.
NQE successively estimates the one dimensional quantiles of each dimension of $\btheta$, conditioned on the data $\bx$ and previous $\btheta$ dimensions.
We interpolate the discrete quantiles with monotonic cubic Hermite splines, adopting specific treatments to account for the tail behavior and potential multimodality of the distribution.
Posterior samples can then be drawn by successively applying inverse transform sampling for each dimension of $\btheta$.
We also develop a post-processing calibration strategy, leading to \textbf{guaranteed unbiased posterior estimation} as long as one provides enough ($\lesssim 10^3$) simulations to accurately calculate the empirical coverage.
To the best of our knowledge, this constitutes the first demonstration that QR-based SBI methods can attain state-of-the-art performance, matching or surpassing the benchmarks set by existing methods.

The structure of this paper is as follows:
In \cref{sec:method}, we introduce the methodology of NQE, along with a alternative definition for Bayesian credible regions and a post-processing calibration scheme to ensure the unbiasedness of the inference results.
In \cref{sec:exp}, we demonstrate that NQE attains state-of-the-art performance across a variety of benchmark problems, together with a realistic application to high dimensional cosmology data.
Subsequently, in \cref{sec:dis}, we discuss related works in the literature and potential avenues for future research.
The results in this paper can be reproduced with the publicly available \texttt{NQE} package
\footnote{\url{https://github.com/h3jia/nqe}.} based on \texttt{pytorch} \citep{paszke2019pytorch}.

\section{Methodology}

\label{sec:method}

\subsection{Quantile Estimation And Interpolation}

\label{sec:2.1}

\begin{figure}[ht]
\vskip 0.2in
\begin{center}
\centerline{\includegraphics[width=0.9\columnwidth]{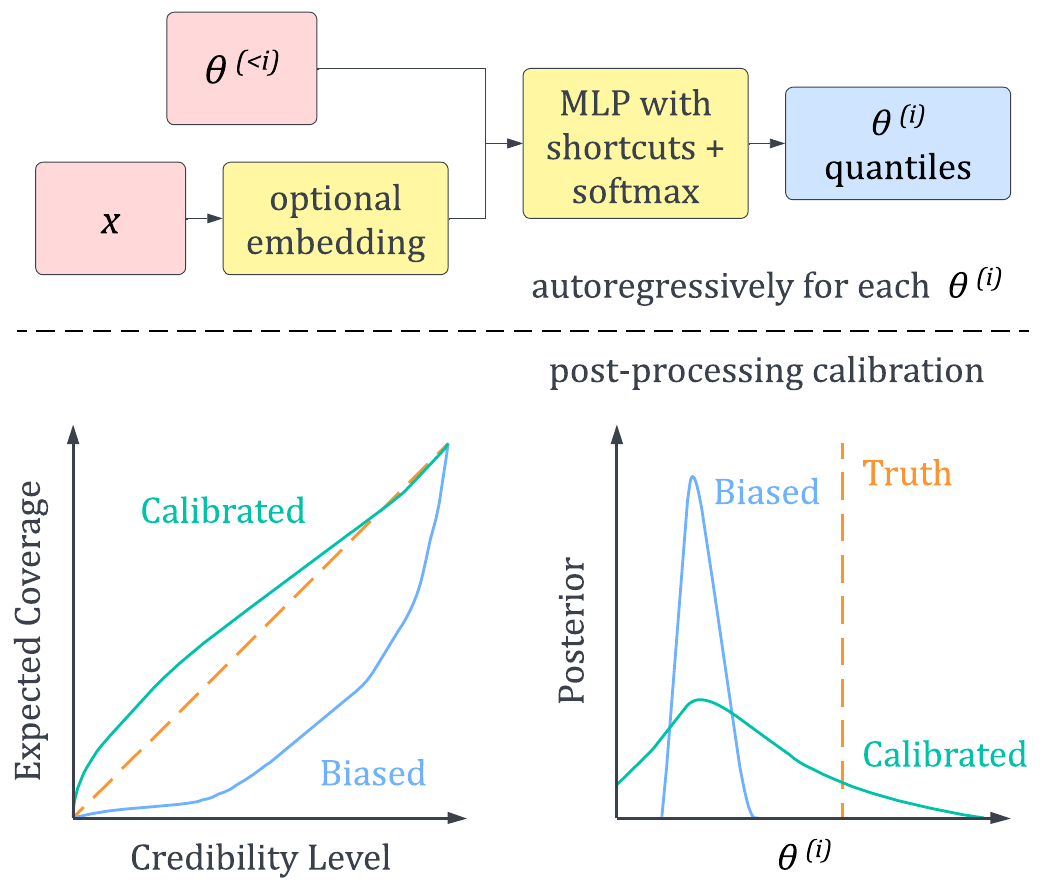}}
\caption{
(Top) Network architecture of our NQE method, which autoregressively learns 1-dim conditional quantiles for each posterior dimension.
The estimated quantiles are then interpolated to reconstruct the full distribution.
(Bottom) A post-processing calibration step can be employed to ensure the unbiasedness of NQE inference results.
\label{fig:nqe}}
\end{center}
\vskip -0.2in
\end{figure}

The cornerstone of most contemporary SBI methods is some form of conditional density estimator, which is used to approximate the likelihood, the posterior, or the likelihood-to-evidence ratio. Essentially, every generative model can function as a density estimator. While Generative Adversarial Networks \citep{goodfellow2020generative} and more recently Diffusion Models \citep{dhariwal2021diffusion} have shown remarkable success in generating high-quality images and videos, the SBI realm is primarily governed by Normalizing Flows \citep[NF, e.g.][]{rezende2015variational,papamakarios2019normalizing}, which offer superior inductive bias for the probabilistic distributions with up to dozens of dimensions frequently encountered in SBI tasks. Our proposed NQE method can also be viewed as a density estimator, as it reconstructs the posterior distribution autoregressively from its 1-dim conditional quantiles.


\begin{figure}[!h]
\vskip 0.2in
\begin{center}
\centerline{\includegraphics[width=0.93\columnwidth]{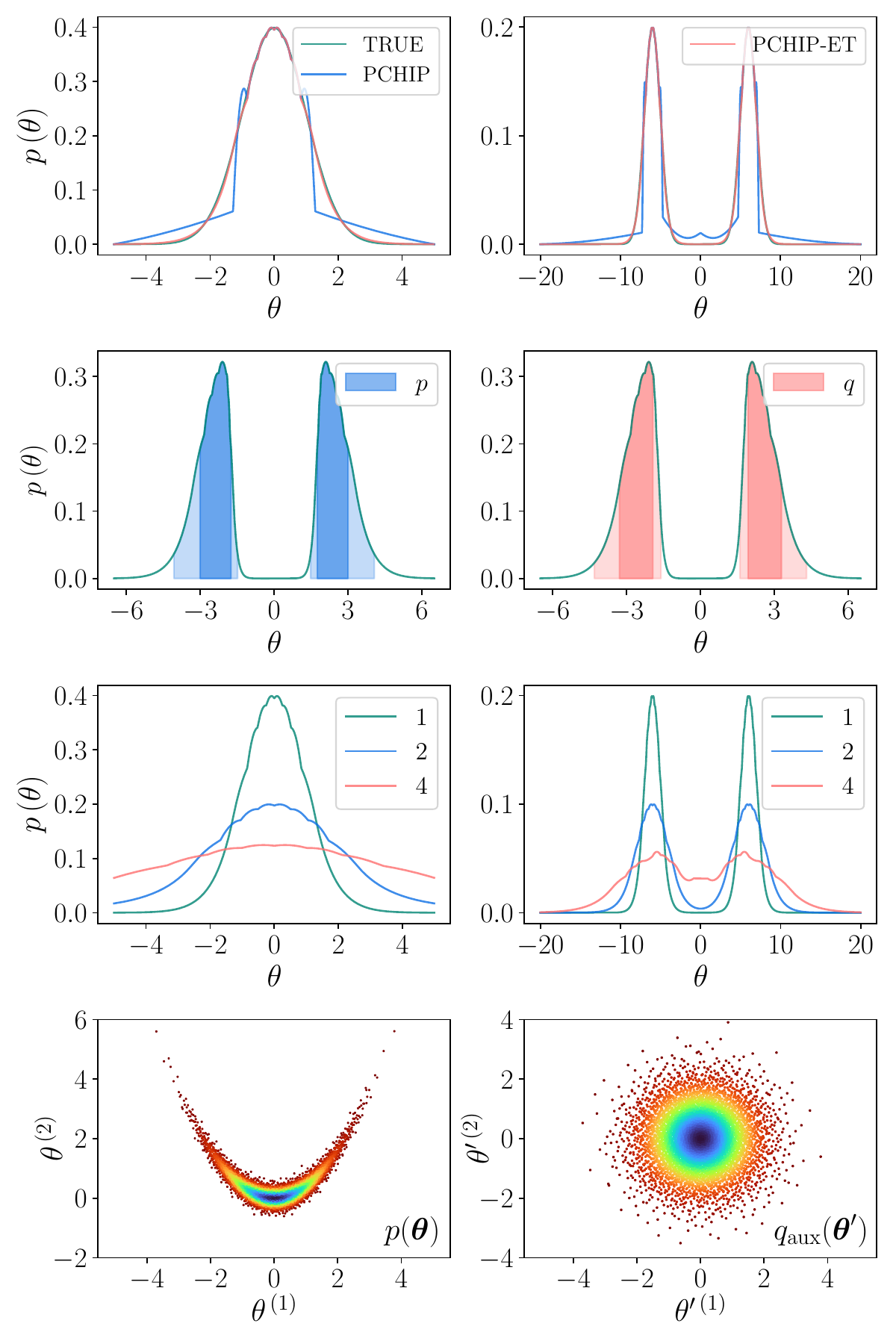}}
\caption{
(1st row) Interpolation of Gaussian and Gaussian Mixture distributions.
While the original PCHIP algorithm shows significant interpolation artifacts, our modified PCHIP-ET scheme decently reconstructs the distributions with only $\sim 15$ quantiles.
(2nd row) Comparison of the 68.3\% and 95.4\% credible regions for a mixture of two asymmetric modes, evaluated with HPDR ($p$-coverage) and QMCR ($q$-coverage).
(3rd row) Broadening of the interpolated posterior, with the broadening factors indicated in the legend.
(4th row) The bijective mapping $f_{\rm qm}$ establishes a one-to-one correspondence between $\btheta$ and $\btheta'$ with the same 1-dim conditional CDF across all the $\theta^{(i)}$ dimensions. The $p-$coverage and $q-$coverage are based on the ranking of $p(\btheta)$ and $q_{\rm aux}(\btheta')$, respectively.
\label{fig:interp}}
\end{center}
\vskip -0.2in
\end{figure}

In a typical SBI setup, one first samples the model parameters $\btheta$ from the prior $p(\btheta)$, and then runs the forward simulations to generate the corresponding observations $\bx$.
For simplicity, let us start with the scenario of 1-dim $\theta$.
Given a dataset $\{\theta,\bx\}$ and a neural network $F_{\phi}(\bx)$ parameterized by $\phi$, one can estimate the median (mean) of $\theta$ conditioned on $\bx$ by minimizing the $L_1$ ($L_2$) loss \footnote{Not to be confused with $\mathcal{L}_0$ and $\mathcal{L}_1$ defined below.} between $\theta$ and $F_{\phi}(\bx)$.
As a straightforward generalization, one can estimate the $\tau$-th quantile of $\theta$ conditioned on $\bx$ by minimizing the following weighted $L_1$ loss,
\begin{align}
    \mathcal{L}_{\tau}[\theta,\,F_{\phi}({\bx})] \, \equiv \ &(\tau-1)\sum_{\theta<F_{\phi}(\bx)} w({\bx})\,\left[\theta-F_{\phi}({\bx})\right]\,+ \notag \\
    &\tau\sum_{\theta\geq F_{\phi}(\bx)} w({\bx}) \left[\theta-F_{\phi}({\bx})\right]\,.\label{eq:ltau}
\end{align}
Here one can introduce an additional $\bx$-dependent weight $w({\bx})$,
similar to the fact that one can use simulations allocated from an arbitrary prior to train SNLE.
A discussion regarding the choice of $w({\bx})$ can be found in \cref{app:wxl0}.
To reconstruct the full posterior, we require the quantiles at multiple $\tau$'s, for which we aggregate the individual loss functions,
\begin{equation}
    \mathcal{L}_0[\theta,\,F_{\phi}({\bx})] \equiv \sum_{\tau} \mathcal{L}_{\tau}[\theta,\,F_{\phi}({\bx})]\,.\label{eq:l0}
\end{equation}

Without loss of generality, we assume the prior of $\theta$ is zero outside some interval $[a, b]$.
If the prior is positive everywhere on $\mathbb{R}$, one can choose $[a, b]$ such that the prior mass outside it is negligible.
For example, one can set $[a, b]$ to $[-5, 5]$ for a standard Gaussian prior; in case of heavy-tailed priors, one can also use the (inverse) prior CDF to map the prior support to $[0, 1]$.
We then equally divide the interval $[0, 1]$ into $n_{\rm bin}$ bins, and estimate the corresponding $n_{\rm bin}-1$ quantiles with $F_{\phi}(\bx)$.
In this work, we choose $F_{\phi}(\bx)$ to be a Multi-Layer Perceptron (MLP) with $n_{\rm bin}$ outputs $z_i$ followed by a softmax layer, such that the $i/n_{\rm bin}$-th quantile of $\theta$ is parameterized as $a+(b-a)\times \sum_{j\leq i} {\rm softmax}(z_j)$, and we add shortcut connections (the input layer of MLP is concatenated to every hidden layer) to facilitate more efficient information propagation throughout the network.
Moreover, an optional embedding network \citep[e.g.][]{jiang2017learning,radev2020bayesflow} can be added before the MLP to more efficiently handle high dimensional data (e.g. the cosmology example in \cref{sec:cosmo}).

For multidimensional $\btheta$, we successively apply the aforementioned method to each dimension $\theta^{(i)}$, conditioned on not only the data $\bx$ but also all the previous dimensions $\btheta^{(j<i)}$.
In other words, $F_{\phi}(\bx)$ in \cref{eq:ltau,eq:l0} is replaced by $F_{\phi}(\bx,\,\btheta^{(j<i)})$, since $\btheta^{(j<i)}$ is effectively treated as observation data for the inference of $\theta^{(i)}$.
An illustration of the NQE architecture can be found in the top panel of \cref{fig:nqe}.
Similar to Flow Matching Posterior Estimation \citep[FMPE,][]{dax2023flow}, NQE has an \textit{unconstrained} architecture which does not require specialized NFs.

\begin{figure}[t]
\vskip 0.2in
\begin{center}
\centerline{\includegraphics[width=0.6\columnwidth]{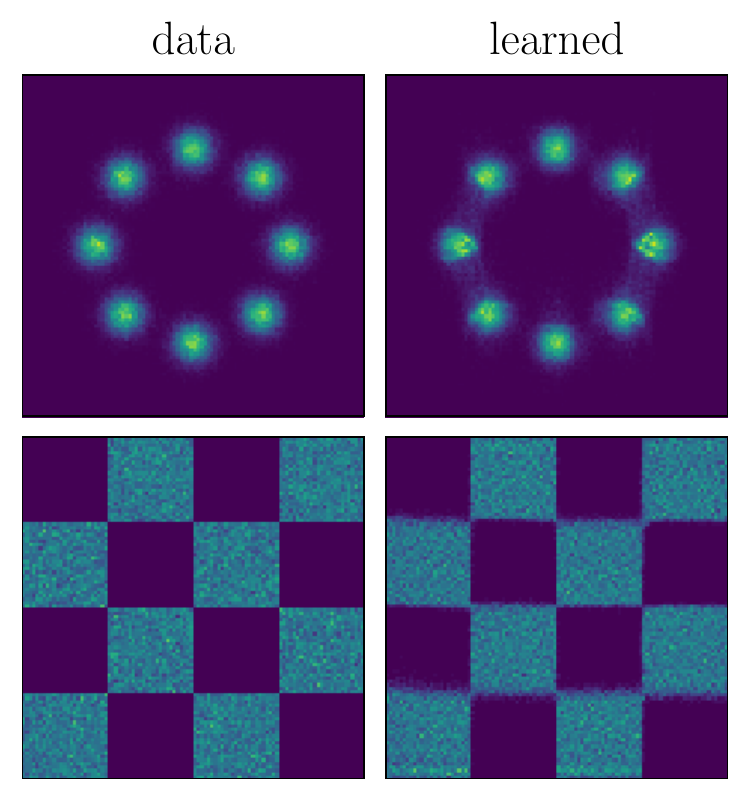}}
\caption{Probability density estimation for two toy examples from \citet{grathwohl2018ffjord}.
Despite the intricate multimodal structures, NQE is able to faithfully reconstruct both distributions.
\label{fig:toy}}
\end{center}
\vskip -0.2in
\end{figure}


The estimated conditional quantiles must be interpolated to enable sampling from them. We achieve this by interpolating the Cumulative Distribution Function (CDF) using Piecewise Cubic Hermite Interpolating Polynomial with Exponential Tails (PCHIP-ET), a modified version of the PCHIP scheme \citep{fritsch1980monotone}, which preserves monotonicity of input data and continuity of first derivatives, ensuring a well-defined Probability Distribution Function (PDF). As depicted in the 1st row of \cref{fig:interp}, the original PCHIP algorithm presents discernible interpolation artifacts, primarily because polynomials cannot decay rapidly enough to align with the true PDF in the tail regime. To address this issue, we substitute the polynomials with Gaussians within bins identified as tails. A more detailed description of our PCHIP-ET scheme is available in \cref{app:interp}. We observe that a satisfactory reconstruction of unimodal distributions can be achieved with $\lesssim 15$ quantiles, while incorporating additional bins may facilitate better convergence in multimodal cases. Samples can then be drawn using inverse transform sampling with the interpolated CDF.



NQE requires
one neural network for each posterior dimension, which can be trained independently on multiple devices to reduce the training wall time.
In principle, one can also train NQE by maximizing the joint PDF, similar to the training of NPE.
However, such approach will be less efficient than minimizing $\mathcal{L}_0$ in \cref{eq:l0}, since one needs to compute the PCHIP-ET interpolation for the PDF, while $\mathcal{L}_0$ only depends on the individual quantiles.
NQE can also be used to estimate $p(\btheta)$ distributions with no observation $\bx$ to condition on.
In this case, we do not need neural networks for the first dimension $\theta^{(1)}$, which can be directly interpolated from the empirical quantiles.
In \cref{fig:toy}, we demonstrate that NQE can successfully model two complicated distributions from \citet{grathwohl2018ffjord}.

\subsection{Regularization}

\label{sec:2.2}

Numerical derivatives are inherently noisier than integrals, and similarly for the PDF compared with the CDF.
To mitigate this issue, we propose the following regularization scheme to improve the smoothness of NQE PDF predictions.
Intuitively, a ``smooth distribution" means the averaged PDF within every 1-dim bin for quantile prediction, $\left<{p}\right>_{\rm bin}$, should be close to the interpolated value between its neighboring bins,
\begin{align}
    \left<{p}\right>_{\rm interp} \equiv \max \biggl[ &f_1\times\left(\left<{p}\right>_{\rm left}+\left<{p}\right>_{\rm right}\right)/\,2,\nonumber \\
    &f_2\times\max\left(\left<{p}\right>_{\rm left},\,\left<{p}\right>_{\rm right}\right) \biggr]\,, \label{eq:pinterp}
\end{align}
with $f_1=1.1$ and $f_2=0.8$, which leads to the following loss for regularization,
\begin{align}
    \mathcal{L}_1 \equiv \sum_{\rm bins}\ &H\left(\left<{p}\right>_{\rm bin} - \left<{p}\right>_{\rm interp}\right) \times \notag \\
    &\left(\log\left<{p}\right>_{\rm bin} - \log\left<{p}\right>_{\rm interp}\right)^2,\label{eq:l1}
\end{align}
where $H(\cdot)$ is the Heaviside function.
With \cref{eq:l1}, we only penalize cases where $\left<{p}\right>_{\rm bin} > \left<{p}\right>_{\rm interp}$, since we will have $\left<{p}\right>_{\rm bin} < \left<{p}\right>_{\rm interp}$ between the peaks in multimodal problems, which is therefore a possible feature in the ground truth solution that should not be penalized.
For similar reasons, $\left<{p}\right>_{\rm interp}$ in \cref{eq:pinterp} is set to be larger than the naive average of $\left<{p}\right>_{\rm left}$ and $\left<{p}\right>_{\rm right}$, so that the regularization is only activated when necessary.
The total loss is then defined as
\begin{equation}
    \mathcal{L} \equiv \mathcal{L}_0\,(1+\lambda_{\rm reg}\mathcal{L}_1)\,. \label{eq:ltotal}
\end{equation}
Note that a linear rescaling of $\btheta$ changes $\mathcal{L}_0$ while $\mathcal{L}_1$ remains invariant, which motivates our choice of $\mathcal{L}$ above.
We find 0.1 to be a generally reasonable choice for $\lambda_{\rm reg}$, although one may reduce $\lambda_{\rm reg}$ for examples with e.g. sharp spikes or edges in the posterior distribution, if one has such prior knowledge of the typical shape of the posterior.



\subsection{Empirical Coverage}

\label{sec:2.3}

Analogous to frequentist confidence regions, Bayesian statistics utilizes credible regions to define the reasonable space for model parameters $\btheta$ given $\bx$.
The most popular choice of Bayesian credible region, namely the highest posterior density region \citep[HPDR, e.g.][]{mcelreath2020statistical}, encloses the $\alpha\%$ samples with the highest PDF for the $\alpha\%$ credible region,
achieving the smallest $\btheta$ volume for any given credibility level.
To test whether a posterior estimator is biased, one checks the empirical coverage, namely the probability of the true model parameters to fall into the $\alpha\%$ credible region over the simulation data.
If such probability is larger (smaller) than $\alpha\%$, the posterior estimator is over-conservative (biased) \footnote{Note that being well calibrated is a necessary yet not sufficient condition for an estimator to predict the Bayesian optimal posterior, as exemplified by the extreme case where the posterior estimator always outputs the prior.}.
To compute the empirical coverage in practice, one needs to pick $N_o$ pairs of $(\btheta,\bx)$ from the simulation data, and generate $N_r$ samples for each of them to get the rank of PDF, leading to $\mathcal{O}(N_oN_r)$ neural network calls for NPE and NQE \footnote{We ignore the factor $\dim \btheta$ for NQE as we define one network call as one evaluation of the whole estimator.}.
For NLE and NRE, such cost is further multiplied by $N_m$, the number of posterior evaluations per effective MCMC sample
\footnote{For $\dim \btheta \lesssim 5$ one may circumvent MCMC using Importance Sampling, which however becomes inefficient as the dimensionality of $\btheta$ grows.}.
Typically one needs to set both $N_o$ and $N_r$ to $\sim10^2-10^3$ so as to get a reliable estimate of the empirical coverage, leading to a moderate computational cost especially for NLE and NRE methods.

A unique characteristics of NQE is that it predicts the distribution quantiles, which explicitly contains the information regarding the \textit{global} properties of the posterior and enables us to propose the following quantile mapping credible region (QMCR) \footnote{Not to be confused with the quantile mapping technique used to e.g. correct the bias for simulated climate data \citep{maraun2013bias}.}, a generalization of the 1-dim equal-tailed credible interval \citep[e.g.][]{mcelreath2020statistical} for multidimensional distributions.
\citet{talts2018validating} shows the rank of any 1-dim statistic can be used to define the Bayesian credible region, with HPDR a special case that chooses such statistic as the posterior PDF.
With the conditional quantiles predicted by NQE, we introduce an auxiliary distribution $q_{\rm aux}(\btheta')$, which we typically set to a multivariate standard Gaussian.
We then define a bijective mapping $f_{\rm qm}:\btheta\to\btheta'$ that establishes a one-to-one correspondence between $\btheta$ and $\btheta'$ with the same 1-dim conditional CDF, $\int p(\theta^{(i)}\,|\,\bx,\,\btheta^{(j<i)})\,{\rm d}\theta^{(i)}$ and $\int q_{\rm aux}(\theta'^{(i)}\,|\,\btheta'^{(j<i)})\,{\rm d}\theta'^{(i)}$, across all the $\theta^{(i)}$ dimensions$\ $\footnote{If $q_{\rm aux}(\btheta')$ is set to a multivariate standard Gaussian, there is no correlation between the different dimensions so we indeed have $q_{\rm aux}(\theta'^{(i)}|\btheta'^{(j<i)})=q_{\rm aux}(\theta'^{(i)})$.}.
The defining statistic of the credible region is chosen as $\log q_{\rm aux}(\btheta')$ with $\btheta'=f_{\rm qm}(\btheta)$, whose rank can be computed analytically using the $\chi^2$ distribution since $q_{\rm aux}(\btheta')$ is Gaussian.
If the interpolation indicates that $p(\theta^{(i)}\,|\,\bx,\,\btheta^{(j<i)})$ includes multiple modes, we use the \textit{local} CDF within the mode containing $\theta$ to define the mapping $f_{\rm qm}$, such that the low PDF regions between the modes are excluded from the credible regions.

A comparison of HPDR and QMCR for a toy distribution can be found in the 2nd row of \cref{fig:interp}, together with the $f_{\rm qm}$ mapping illustrated in the 4th row.
Heuristically, the $\alpha\to0$ limit of QMCR encloses the (conditional) median across all the dimensions for unimodal distributions, as opposed to the global maximum of the PDF for HPDR.
Therefore, unlike HPDR, QMCR is invariant under any 1-dim monotonic transforms of $\btheta$, as long as such reparameterization does not give rise to a different identification of multimodality during the CDF interpolation.
As shown with the examples below, QMCR typically leads to similar conclusions regarding the (un)biasedness of the posterior estimators as HPDR, but only requires $\mathcal{O}(N_o)$ network calls to evaluate as one no longer needs to generate $N_r$ samples for each observation.
Such speed-up allows us to perform posterior calibration in the next subsection with negligible computational cost.
For simplicity, in the rest of this paper we will use the term \textbf{$\bm p-$coverage ($\bm q-$coverage)} for empirical coverage computed with HPDR (QMCR).
In addition, we note that due to its autoregressive structure, one can compute the coverage of NQE for the leading $\btheta$ dimensions without additional training, which is useful if the unbiasedness of certain $\btheta$ dimensions takes precedence over others.

\subsection{Posterior Calibration}

\label{sec:2.4}




\citet{hermans2021trust} demonstrates that all existing SBI methods may produce biased results when the simulation budget is limited.
Intuitively, a biased posterior is too narrow to enclose the true model parameters, so we propose the following calibration strategy as illustrated in the bottom panel of \cref{fig:nqe}.
To make a distribution broader, we fix the medians of all 1-dim conditional posteriors and increase the distance between the medians and all other quantiles by a global \textit{broadening factor}.
Similar to the $q-$coverage evaluation, we utilize the local quantiles within modes for multimodal distributions.
We remove the quantiles that escape from the boundary of the prior and/or the boundary between different modes, and redistribute the corresponding posterior mass to the bins still within the boundary based on the bin mass, so that the local posterior shape is preserved.
The effect of such broadening transform is shown in the 3rd row of \cref{fig:interp}.
We then solve for the \textit{minimum broadening factor} such that the calibrated posterior is unbiased  across a series of credibility levels, which we set to $\{0.1,\,0.5,\,0.9\}$ throughout this paper.
Note that ideally, a good estimator should have empirical coverage that matches the credibility level.
However, if this is not possible due to limited training data, over-conservative inference should be preferred over biased results.
The broadening factor can also be smaller than 1, in case the original posterior is already too conservative.
While one has the freedom to choose the definition of the coverage for the calibration process, the broadened posterior is only guaranteed to be unbiased at the calibrated credibility levels under the same coverage definition.

\begin{table}[t]
\caption{Computational cost of the broadening calibration, with NQE being significantly faster than other methods. $N_i$: number of iterations to solve for the desired coverage. $N_o$: number of simulated observations for coverage computation. $N_r$: number of samples per observation for the rank of PDF. $N_m$: number of posterior evaluations per effective MCMC sample. We assume there is no broadening technique for NPE that does not necessitate MCMC sampling.} \label{tab:cali}
\begin{center}
\begin{tabular}{cccc}
 & coverage & simulations & network calls \\
\hline \vspace{-0.3cm} \\
\bf{NQE} & $\bm q$ & $\bm N_o$ & $\bm{\mathcal{O}( N_o )}$ \\
NQE & $p$ & $N_o$ & $\mathcal{O}( N_i N_o N_r )$ \\
NLE & $p$ & $N_o$ & $\mathcal{O}( N_i N_o N_r N_m )$ \\
NPE & $p$ & $N_o$ & $\mathcal{O}( N_i N_o N_r N_m )$ \\
NRE & $p$ & $N_o$ & $\mathcal{O}( N_i N_o N_r N_m )$ \\
\end{tabular}
\end{center}
\end{table}

While similar calibration tricks may also be developed for other SBI methods, it will likely be considerably more expensive than NQE in practice, for the following reasons.
Firstly, the evaluation of $q-$coverage is exclusive to NQE, which is faster by at least a factor of $N_r$ than traditional $p-$coverage (with an additional factor of $N_m$ if MCMC is required for sampling).
More importantly, we have developed a broadening strategy for NQE that preserves not only the local correlation structure of the posterior but also the ability of fast sampling without MCMC.
We are not aware of any similar techniques for existing SBI methods, which estimate the local PDF with no explicit global information of the distribution.
For example, while one can broaden an NF-based probability distribution by lowering its \textit{temperature}, i.e. replacing $\log p(\btheta|\bx)$ with $\beta\log p(\btheta|\bx)$, $\beta<1$, this will necessitate MCMC sampling for NPE (NLE and NRE need MCMC even without broadening).
In addition, with the analytical rank evaluation of $q-$coverage, the NQE network outputs can also be reused between different iterations, thus reducing the total network calls by another factor of $N_i$.
We compare the computational cost of broadening calibration for different methods in \cref{tab:cali}.

Such post-processing calibration relies on a reliable calculation of the coverage.
The (pointwise) error of empirical coverage due to stochastic sampling can be estimated using binomial distribution \citep{sailynoja2022graphical}; with $N_o=10^3$, the maximum error is smaller than $1.6\%$, regardless of the dimensionality of $\bx$ and $\btheta$ \footnote{See \cref{app:cover} for more discussion on this.}.
In other words, \textbf{for any inference task, with the broadening calibration, one only needs $\bf \lesssim 10^3$ simulations in the validation dataset to ensure the unbiasedness of the posterior, if there is no model misspecification}.
Nevertheless, the number of network calls required for broadening is different across the various algorithms as compared in \cref{tab:cali}.
Using NQE and $q-$coverage, one only needs $\mathcal{O}(N_o)$ calls of the NQE network for the broadening, which is typically negligible compared with the cost for running the simulations and training the neural estimators.
In addition, \textbf{similar calibration tricks can be used to mitigate partially known model misspecification, as exemplified in \cref{sec:cosmo} below.}
Note that we use the same validation dataset during the training and broadening calibration of NQE, as the one-parameter broadening transform is unlikely to overfit.
We summarize the proposed NQE method in \cref{alg:nqe}.

\begin{algorithm}[tb]
   \caption{Neural Quantile Estimation (NQE)}
   \label{alg:nqe}
\begin{algorithmic}
    \STATE \hspace*{-0.45cm} {\bfseries 1.$\ $Training NQE} \hfill $\blacktriangleright$ \cref{sec:2.1,sec:2.2}
    \STATE \hspace*{0.025cm} {\bfseries for} $i=1$ to $\dim\theta$ {\bfseries do} \hfill $\blacktriangleright$ parallelizable
    \STATE \hspace*{0.35cm} train $F_{\phi}(\bx,\,\btheta^{(j<i)})$ by minimizing $\mathcal{L}$ \hfill $\blacktriangleright$ \cref{eq:ltotal}
    \STATE \hspace*{-0.45cm} {\bfseries 2.$\ $Calibrating NQE (Optional)} \hfill $\blacktriangleright$ \cref{sec:2.3,sec:2.4}
    \STATE \hspace*{0.025cm} solve the calibration for unbiased posterior
    \STATE \hspace*{-0.45cm} {\bfseries 3.$\ $Sampling NQE} \hfill $\blacktriangleright$ \cref{sec:2.1}
    \STATE \hspace*{0.025cm} {\bfseries for} $i=1$ to $\dim\theta$ {\bfseries do} \hfill $\blacktriangleright$ sequential
    \STATE \hspace*{0.35cm} sample $\theta^{(i)}$ from interpolated $F_{\phi}(\bx,\,\btheta^{(j<i)})$
\end{algorithmic}
\end{algorithm}

In this paper, we focus on the simple broadening calibration, which is guaranteed to converge with $\lesssim 10^3$ validation simulations, regardless of $\dim \bx$ and $\dim \btheta$.
With more simulations, it may be beneficial to employ a more sophisticated calibration scheme to remove the bias without over-broadening the predicted posterior.
We plan to conduct a comprehensive survey of such calibration schemes in a follow-up paper.
One example is the \textit{quantile shifting} calibration demonstrated with the cosmology example in \cref{sec:cosmo}: for each $\tau$ quantile of $p(\theta^{(i)} | \bx, \btheta^{(<i)})$ predicted by NN, we check if we indeed have $\tau$ probability that the true $\theta^{(i)}$ is smaller than the predicted quantile (on the validation dataset) \footnote{For multi-modal distributions, we use the local quantile within the mode that contains the true $\theta^{(i)}$, similar to the definition of the $q-$coverage in \cref{sec:2.3}.}.
If not, we calculate the shift required for the $\tau$ quantile such that this statement is true.
Note that we apply a shift of $\theta^{(i)}$ quantile that is different for each $\tau$ and $i$, but the same for all $\bx$ and $\btheta^{(<i)}$.
In other words, we effectively calculate the bias averaged over the prior, and shift the predicted quantiles accordingly to remove the bias.
Strictly speaking, such quantile shifting scheme calibrates the $q-$coverage of all the individual 1-dim conditional posteriors, but not necessarily the $q-$coverage of the multi-dimensional joint posterior.
In addition, the number of simulations required for this scheme depends on the dimensionality of $\btheta$, in contrast to the global broadening scheme which always converges with $\lesssim10^3$ validation simulations.
We leave a more detailed investigation of such methods for future research; nevertheless, for the cosmology example in \cref{sec:cosmo}, the posterior calibrated with quantile shifting has an almost diagonal empirical coverage and is much narrower than the posterior calibrated with simple global broadening, when there is a significant bias in the uncalibrated posterior due to model misspecification.

\begin{figure}[t]
\vskip 0.2in
\begin{center}
\centerline{\includegraphics[width=\columnwidth]{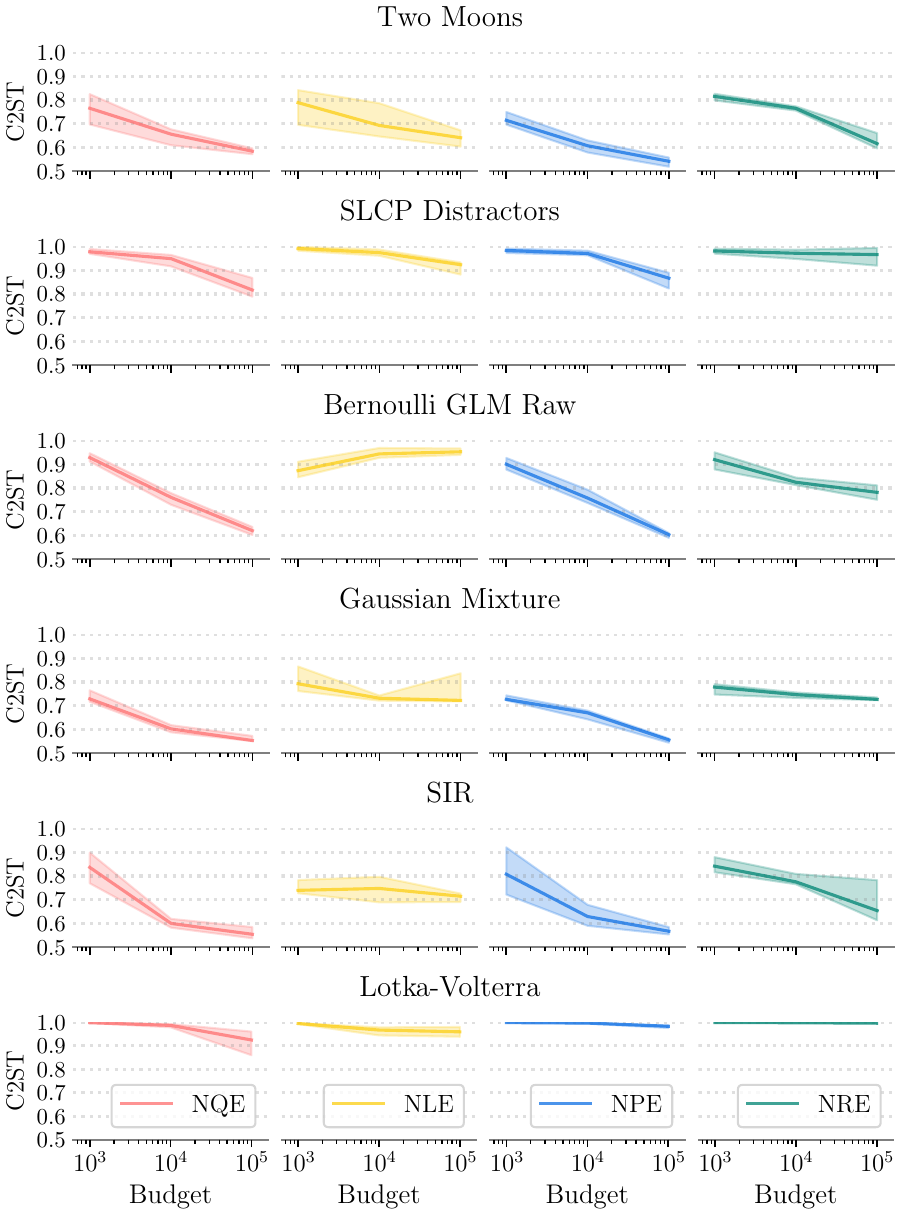}}
\caption{Comparison of C2ST as a function of simulation budget for the six benchmark problems, with lower C2ST values representing better performance of the algorithm.
The error bars are estimated using the 25\%, 50\% and 75\% quantiles of C2ST over ten realizations for each problem.
(Uncalibrated) NQE achieves state-of-the-art performance across all the examples.\label{fig:02-ns}}
\end{center}
\vskip -0.2in
\end{figure}

\begin{figure*}[ht]
\vskip 0.2in
\begin{center}
\centerline{\includegraphics[width=0.95\textwidth]{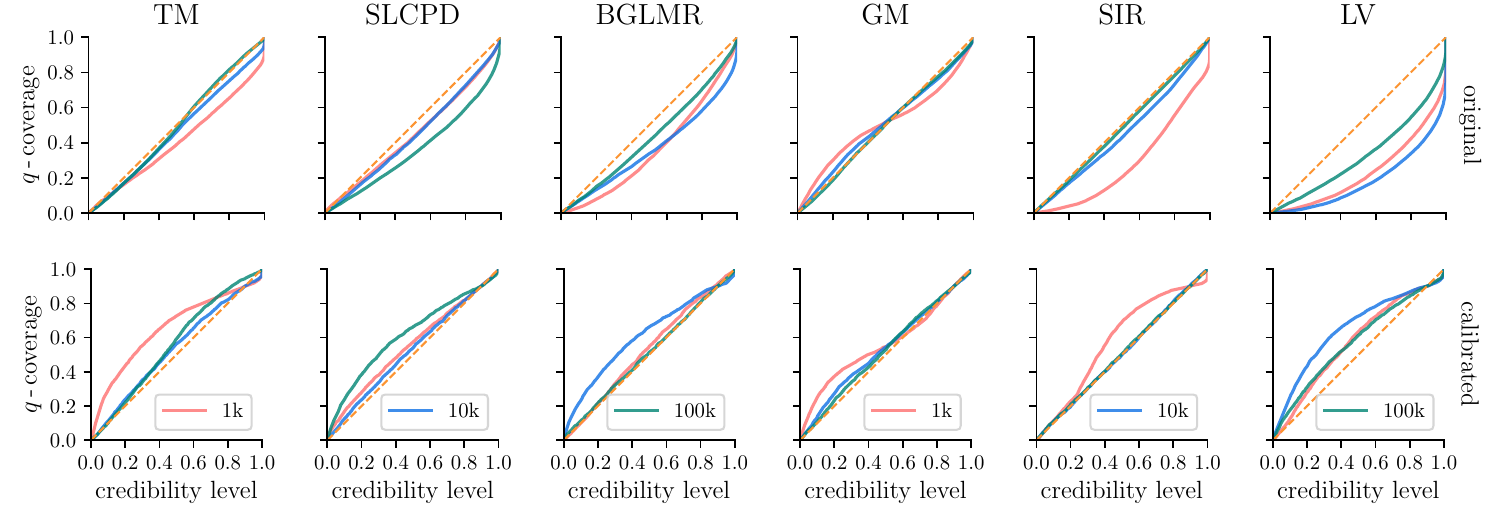}}
\caption{
(Top) NQE $q-$coverage for the benchmark problems. Like other SBI methods, with limited simulation budgets, NQE may predict biased posteriors.
(Bottom) Calibrated NQE predicts unbiased posteriors for all the problems.
Errorbars are small and thus not plotted.
See \cref{app:cover} for a convergence test and \cref{fig:p-cover} for a similar plot with $p-$coverage.
\label{fig:q-cover}}
\end{center}
\vskip -0.2in
\end{figure*}

\begin{figure*}[ht]
\vskip 0.2in
\begin{center}
\centerline{\includegraphics[width=0.85\textwidth]{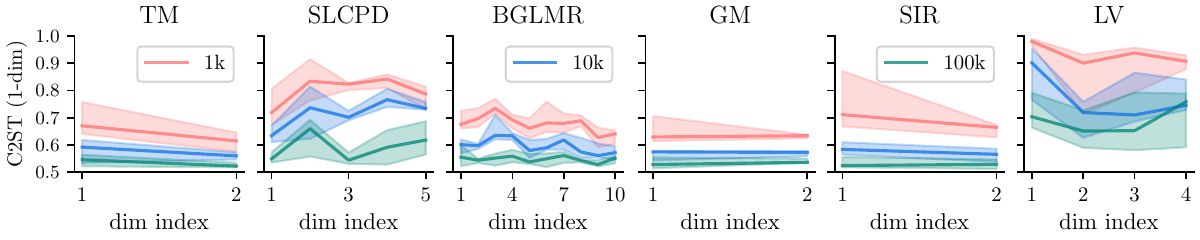}}
\caption{
The C2ST values for the 1-dim \textit{uncalibrated} marginal posteriors. We do not observe a clear trend of increasing C2ST with respect to the ordering of the dimensions.
\label{fig:od}}
\end{center}
\vskip -0.2in
\end{figure*}

\section{Numerical Experiments}

\label{sec:exp}


\subsection{SBI Benchmark Problems}

We assess the performance of NQE on six benchmark problems,
with detailed specifications provided in \cref{app:prob}.
All results for methods other than NQE are adopted from \citet{lueckmann2021benchmarking}.
As discussed in \cref{app:hyper}, we conduct a mild search of hyperparameters for NQE, but in the end use the same set of hyperparameters across all the benchmark problems,
although it is possible to further improve the performance by tuning the hyperparameters based on specific posterior structures.
For example, increasing the number of predicted quantiles will be beneficial for multimodal problems with large simulation budgets.
To evaluate the performance of SBI algorithms, we employ Classifier-based 2-Sample Testing (C2ST) as implemented in the \texttt{sbibm} package \citep{lopez2016revisiting,lueckmann2021benchmarking}. Lower C2ST values denote superior results, with 0.5 signifying a perfect posterior and 1.0 indicating complete failure.

We plot the C2ST results for the benchmark problems in \cref{fig:02-ns}, showing that (uncalibrated) NQE achieves state-of-the-art performance across all the examples.
In \cref{fig:q-cover}, we compare the NQE $q-$coverage before and after broadening: with the broadening calibration, NQE consistently predicts unbiased posterior for all the problems.
While \cref{fig:q-cover} utilizes $10^4$ simulations to enhance the smoothness of the coverage curves, a convergence test in \cref{app:cover} shows that $\lesssim 10^3$ simulations are sufficient for most cases.
The exact values of the broadening factor can be found in \cref{fig:bf}.
In \cref{fig:c2st-b}, we find that the C2ST is generally similar or slightly worse after the global broadening calibration: this is likely due to the nature of the C2ST metric, since a conservative posterior will be similarly penalized as a biased posterior, although the former should be preferred over the latter for most scientific applications \citep[e.g.][]{hermans2021trust,delaunoy2022towards}.



\subsection{Order of Model Parameters}

Due to its autoregressive structure, NQE's performance may be affected by the order of $\btheta$ dimensions.
While each 1-dim conditional distribution $p(\theta^{(i)}\,|\,\bx,\,\btheta^{(j<i)})$ is estimated independently, the 1-dim marginal posterior $p(\theta^{(i)}|\bx)$ does depend on the estimation for all the previous $\btheta^{(j<i)}$ that are correlated with $\theta^{(i)}$, therefore one may expect the marginals for the latter dimensions to be less accurate than the former dimensions as the error will accumulate.
To study this effect, we compute all the 1-dim marginal C2ST's for the benchmark problems and plot them with respect to the dimension indices in \cref{fig:od}.
Contrary to the conjecture above, we find no clear dependence between the marginal C2ST and the dimension index.
Nevertheless, this may be due to the relative low posterior dimensionality of the benchmark problems, such that the accumulation of per-dimension error has not become the dominant contribution.
We still recommend ordering the $\btheta$ dimensions based on the relative importance of the parameters, especially for applications to higher ($\gtrsim10$) dimensional posteriors.
We note that similar to the TMNRE approach \citep{miller2021truncated}, one may estimate the individual marginal posteriors with NQE, if the high dimensionality makes it impractical to accurately model the joint posterior.

\subsection{Application to Cosmology}

\label{sec:cosmo}

\begin{figure}[t]
\vskip 0.2in
\begin{center}
\centerline{\includegraphics[width=\columnwidth]{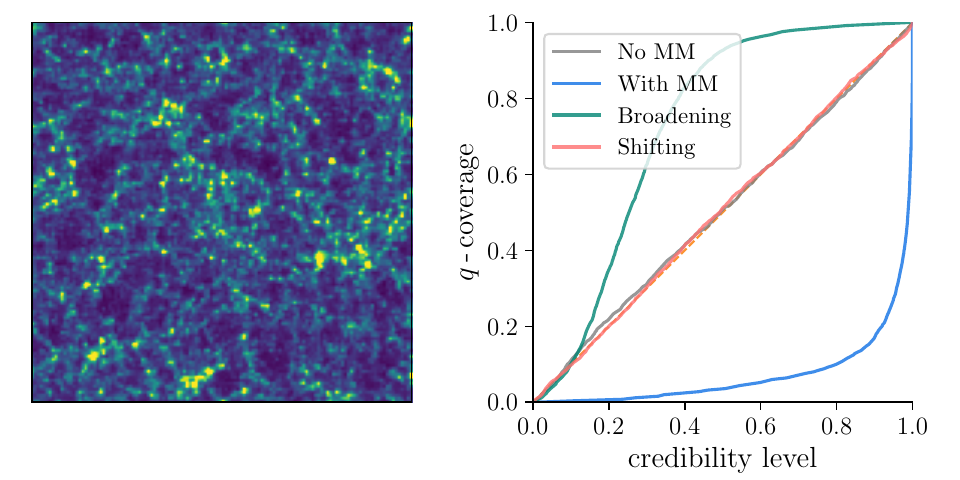}}
\caption{
(Left) Sample image of the simulated data. The task is to infer two parameters of our Universe, $\Omega_m$ and $\sigma_8$, from such images.
(Right) The $q-$coverage for uncalibrated NQE without model misspecification (No MM), uncalibrated NQE with model misspecification (With MM), and NQE with model misspecification but calibrated using a broadening factor of 4.2 (Broadening) and using the quantile shifting method (Shifting).
Both calibration methods eliminate the bias due to \textit{known} model misspecification, with quantile shifting achieving almost exact empirical coverage whereas global broadening being over-conservative.
\label{fig:wl}}
\end{center}
\vskip -0.2in
\end{figure}

\begin{figure}[t]
\vskip 0.2in
\begin{center}
\centerline{\includegraphics[width=\columnwidth]{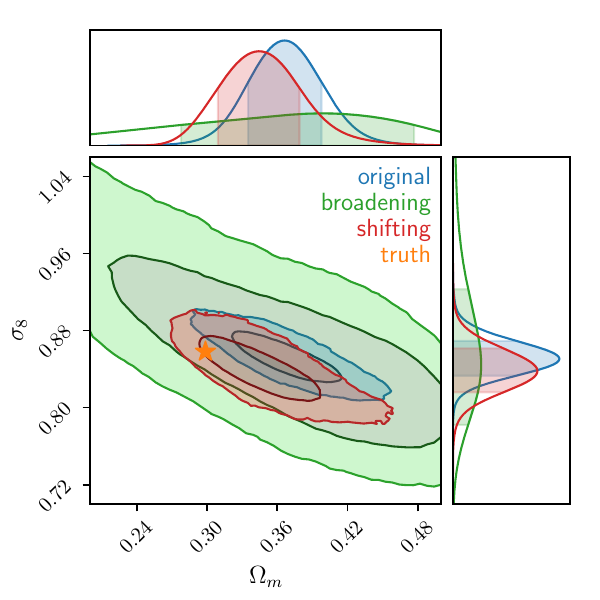}}
\caption{
Comparison of the uncalibrated posterior and the posteriors calibrated with two different schemes.
The quantile shifting scheme removes the bias without over-broadening the posterior.
\label{fig:wl-c}}
\end{center}
\vskip -0.2in
\end{figure}

The cosmological large scale structures contain ample information regarding the origin and future of our universe, which can be inferred from the locations and/or shapes of the galaxies \citep[e.g.][]{dodelson2020modern}, however the optimal strategy to extract the information remains an unsolved problem.
While at larger scales the power spectra carry most of the information and can be well modeled with a Gaussian likelihood, at smaller scales the highly nonlinear evolution render SBI methods necessary for the optimal inference.

Unfortunately, the small-scale baryonic physics is still poorly understood, leading to potential model misspecification which can bias the SBI inference \citep[e.g.][]{modi2023sensitivity}.
As we do not know the exact forward model for our Universe, the best we can do is to make sure our SBI estimator is unbiased on all the well-motivated baryonic physics models, which requires a massive amount of expensive cosmological hydrodynamic simulations \citep[e.g.][]{villaescusa2021camels}.
However, with NQE one can first train it using cheap (therefore less realistic) simulations and then calibrate it using all available high fidelity (therefore much more expensive) simulations to make sure the uncertainties of baryonic physics have been properly accounted for \footnote{Here we assume the model misspecification is at least \textit{partially known}, in the sense that our selection of baryonic physics models ``includes" the correct model for our Universe. The post-processing calibration cannot mitigate \textit{completely unknown} model misspecification.}.
Note that one only needs $\lesssim 10^3$ simulations for each baryonic model to calibrate NQE, which is far fewer than the amount required to directly train field-level SBI with them.
Such approach is demonstrated in \cref{fig:wl,fig:wl-c}, where we show that the bias due to model misspecification can be mitigated by the calibration of NQE.
As the model misspecification introduces a large systematic bias, we find that the global broadening calibration makes the posterior over-conservative while the quantile shifting scheme eliminates the bias without over-broadening the posterior, highlighting the benefits of such more advanced calibration methods that will be examined more thoroughly in a follow-up paper.
More details regarding this example can be found in \cref{app:cosmo}.


\section{Discussion}

The main contribution of this work is to introduce Neural Quantile Estimation (NQE), a novel class of SBI methods that incorporate the concept of quantile regression, with competitive performance across various examples.
Strictly speaking, our paper presents Neural Quantile \textit{Posterior} Estimation, a method that can be extended to Neural Quantile \textit{Likelihood} Estimation, which fits the likelihood $p(\bx|\btheta)$ with conditional quantiles.
We note that the idea of interpolating predicted quantiles has been explored for e.g. time series forecasting \citep{gasthaus2019probabilistic,sun2023neural}.
Nonetheless, to our knowledge our paper is the first work that implements this idea in the SBI framework, with a dedicated interpolation scheme that minimizes the potential artifacts.
In addition, \citet{jeffrey2020solving} uses a similar architecture to predict the moments of the posterior.
\citet{montel2023scalable} proposes to autoregressively apply marginal NRE estimators to obtain the joint distribution, which outperforms standard NRE in their benchmarks.

As shown in \citet{hermans2021trust}, all existing SBI methods may predict biased results in practice: while the Bayesian optimal posterior has perfect calibration, there is no guarantee regarding the unbiasedness of SBI algorithms trained with insufficient number of simulations.
However, with the post-processing calibration step, \textbf{NQE is guaranteed to be unbiased should there be no unknown model misspecification}, in the sense that the credible regions of the posterior will enclose no fewer samples samples than their corresponding credibility levels, as long as one has $\lesssim 10^3$ validation data to reliably compute the empirical coverage for the broadening calibration.
While Balanced Neural Ratio Estimation \citep[BNRE,][]{delaunoy2022towards} pursues similar goals of robust SBI inference, the unbiasedness of BNRE depends on the choice of their regularization parameter, so in principle they need to tune this parameter for each task to obtain best results.
Unfortunately, the coverage evaluation is considerably more expensive for NRE methods which relies on MCMC sampling, making the coverage-based
tuning of BNRE computationally prohibitive for higher dimensional applications.
On the other hand, the broadening calibration of NQE can be applied with negligible computational cost, with the calibrated NQE manifestly unbiased as the empirical coverage has been explictly corrected during the broadening process.
In addition, one can also mitigate the bias due to \textit{partially known} model misspecification by calibrating the NQE posterior.

Before concluding this paper, we enumerate several promising directions for future study.
First of all, NQE can be straightforwardly generalized to Sequential NQE (SNQE), which will be presented in a separate paper.
Second, while our PCHIP-ET scheme shows competitive performance across various problems, it does not have continuous PDF derivatives, which may be improved by a higher order interpolation scheme.
Moreover, in this work we mostly restrict to a global broadening transform for the calibration of NQE, which eliminates the bias at the cost of being possibly too conservative for certain credibility levels.
As shown in \cref{sec:cosmo}, a more advanced calibration strategy would be useful, in particular for problems with a large systematic bias, so that one can calibrate biased posteriors without losing too much constraining power.

\label{sec:dis}

\section*{Acknowledgements}
We thank Sihao Cheng, Biwei Dai, ChangHoon Hahn, Francois Lanusse, Jiaxuan Li, Yin Li, Peter Melchior, Chirag Modi, Nikhil Padmanabhan, Oliver Philcox and David Spergel for helpful discussions.
The work presented in this article was performed on computational resources managed and supported by Princeton Research Computing, a consortium of groups including the Princeton Institute for Computational Science and Engineering (PICSciE) and the Office of Information Technology's High Performance Computing Center and Visualization Laboratory at Princeton University.

\section*{Impact Statement}

This paper presents work whose goal is to advance the field of Machine Learning. There are many potential societal consequences of our work, none which we feel must be specifically highlighted here.

\section*{References}
\vspace{-0.6cm}
\renewcommand{\refname}{}

\bibliography{refs}
\bibliographystyle{icml2024}

\newpage
\appendix
\onecolumn

\section{Piecewise Cubic Hermite Interpolating Polynomial with Exponential Tails}

\label{app:interp}

We interpolate the CDF of the conditional 1-dim distributions using the quantiles predicted by NQE.
Our interpolation scheme is based on Piecewise Cubic Hermite Interpolating Polynomial \citep[PCHIP,][]{fritsch1980monotone,moler2004numerical}, which preserves the monotonicity of the input data and has continuous first order derivatives.
The values of the interpolated function at the $k-$th and $(k+1)-$th nodes, $y_k$ and $y_{k+1}$, match the values of the target function, while the derivatives, $y'_k$ and $y'_{k+1}$, are given by the two-side scheme for non-boundary points,
\begin{align}
    h_k &\equiv x_{k+1} - x_k,\quad d_k \equiv (y_{k+1}-y_k)/h_k,\nonumber \\
    y'_k &= \begin{dcases}
        0,\quad d_k d_{k+1} \leq 0 \\
        \dfrac{w_1+w_2}{w_1 / d_{k-1} + w_2 / d_k},\quad d_k d_{k+1} > 0,\quad\text{where}\ \ w_1 = 2h_k+h_{k-1},\ \ w_2 = h_k+2h_{k-1}.
        \end{dcases}
    \label{eq:2side}
\end{align}
For boundary points, we use the following one-side scheme for the left end (similarly for the right end),
\begin{align}
    y'_1 = \dfrac{(2h_1+h_2)d_1-h_1d_2}{h_1+h_2},
    \label{eq:1side}
\end{align}
which however is clipped to $[p_{\rm tl}d_1,\ 3d_1]$ for $d_1\geq0$ and $[3d_1,\ p_{\rm tl}d_1]$ for $d_1<0$, with $p_{\rm tl}$ a hyperparameter typically set to $0.6$.
Note that for well-defined CDF data, one always has $d_k>0$ in \cref{eq:2side,eq:1side}.
\citet{fritsch1980monotone} shows a sufficient condition for the interpolation to preserve monotonicity is $0 \leq y'_k/d_k \leq 3$ and $0 \leq y'_{k+1}/d_k \leq 3$ \footnote{Indeed 3 is the largest number for the criterion of this form.}, which is satisfied by \cref{eq:2side,eq:1side}.

With $y_k$, $y_{k+1}$, $y'_k$ and $y'_{k+1}$, the interpolation gives
\begin{align} 
    y_{\rm interp}(x)&=h_{00}(t)\times y_k + h_{10}(t)\times h_ky'_k + h_{01}(t)\times y_{k+1} + h_{11}(t)\times h_ky'_{k+1},\quad{\rm where} \\
    h_{00}(t)&=2t^3-3t^2+1, \nonumber \\
    h_{10}(t)&=t^3-2t^2+t, \nonumber \\
    h_{01}(t)&=-2t^3+3t^2, \nonumber \\
    h_{11}(t)&=t^3-t^2, \nonumber \\
    t &\equiv (x-x_k) / (x_{k+1} - x_k). \nonumber
\end{align}
As shown in \cref{fig:interp}, this interpolation scheme generates notable artifacts in the PDF, due to the challenge posed by fitting polynomials to the exponentially declining tail of the probability density.

In response to this challenge, we fit the local distribution with Gaussian tails whenever necessary.
In this regime, the fitting PDF is given by
\begin{equation}
    p(x)=p_0\,e^{a(x-x_0)^2+\frac{p'_0}{p_0}(x-x_0)},
    \label{eq:p(x)}
\end{equation}
with $p(x_0)=p_0$ and its first derivative $p'(x_0)=p'_0$ continuous at the end point of the bin.
We then solve the $a$ parameter by requiring that the PDF has correct normalization within the bin, which  can be computed via the following indefinite integrals.
For $p'_0\neq0$, we have
\begin{equation}
    \int p\,{\rm d}x = \begin{dcases}
        \frac{\sqrt{\pi}}{2\sqrt{|a|}}\, p_0\,e^{-\frac{p'^2_0}{4|a|p_0^2}}\times{\rm erfi} \left[ \frac{2|a|p_0(x-x_0)+p'_0}{2\sqrt{|a|}p_0} \right] + C,\quad & a>0 \\
        \frac{p^2_0\,e^{\frac{p'_0}{p_0}(x-x_0)}}{p'_0} + C,\quad & a=0 \\
        \frac{\sqrt{\pi}}{2\sqrt{|a|}}\,p_0\,e^{\frac{p'^2_0}{4|a|p_0^2}}\times{\rm erf} \left[ \frac{2|a|p_0(x-x_0)-p'_0}{2\sqrt{|a|}p_0}\right] + C,\quad & a<0
    \end{dcases}
\end{equation}
while for $p'_0=0$,
\begin{equation}
    \int p\,{\rm d}x = \begin{dcases}
        \frac{\sqrt{\pi}}{2\sqrt{|a|}}\, p_0\times{\rm erfi} \left[ \sqrt{|a|} (x-x_0) \right] +C,\quad & a>0 \\
        p_0(x-x_0)+C,\quad & a=0 \\
        \frac{\sqrt{\pi}}{2\sqrt{|a|}}\, p_0\times{\rm erf} \left[ \sqrt{|a|}(x-x_0) \right] +C,\quad & a<0
    \end{dcases}
\end{equation}
where ${\rm erf}(\cdot)$ and ${\rm erfi}(\cdot)$ are the error function and imaginary error function, respectively.
For $a\neq0$ and $p'_0\neq0$, we use the following expressions which are analytically equivalent but numerically more stable,
\begin{equation}
    \int_0^h p\,{\rm d}x = \begin{dcases}
        \frac{p_0}{\sqrt{|a|}}\left\{ e^{|a|h^2+p'_0h/p_0} D\left[  \frac{2|a|p_0 h+p'_0}{2\sqrt{|a|}p_0} \right] - D\left[ \frac{p'_0}{2\sqrt{|a|}p_0} \right] \right\},\quad a>0\\
        \frac{\sqrt{\pi}\,p_0}{2\sqrt{|a|}} \left\{ e^{-|a|h^2+p'_0h/p_0} {\rm erfcx} \left[ \frac{p'_0-2|a|p_0h}{2\sqrt{|a|}p_0} \right] - {\rm erfcx} \left[ \frac{p'_0}{2\sqrt{|a|}p_0} \right] \right\},\quad a<0
    \end{dcases}
\end{equation}
where $D(\cdot)$ is Dawson's integral and ${\rm erfcx}(\cdot)$ is the scaled complementary error function.
Nonetheless, in rare cases where $a<0$ we set $a=0$ and give up the continuity condition for the first derivative of PDF, and instead solve $p'_0$ for the correct normalization within the bin.

Our criterion for deciding whether a bin should be fitted with exponential tails is as follows.
First of all, the leftmost and rightmost bins have one-sided exponential tails as long as their averaged PDF is smaller than 0.6 times the averaged PDF in the bin next to them, otherwise the edge bins likely have a hard truncation by the prior and are therefore fitted with polynomials.
In addition, we also allow other bins to have double, i.e. $\,p_{\rm exp}^{\rm (left)}$ from left endpoint $x_k$ towards right and $p_{\rm exp}^{\rm (right)}$ from right endpoint $x_{k+1}$ towards left, exponential tails to account for potential multimodality.
For each bin $[x_k,x_{k+1}]$, we attempt to fit the distribution with double exponential tails, and compute
\begin{equation}
    f_{\rm split} \equiv \max[\,p_{\rm exp}^{\rm (left)}(x_{k+1})\,/\,p_{0}(x_{k+1}),\,p_{\rm exp}^{\rm (right)}(x_{k})\,/\,p_{0}(x_{k})\,].
\end{equation}
Note that the PDF is no longer strictly continuous at the bin endpoints when fitted with double exponential tails, and $f_{\rm split}$ quantifies such discontinuity.
We then switch to double exponentials only for bins with local minimum $f_{\rm split}<0.01$, and stick with the PCHIP polynomials for the remaining bins.
The rationale behind this is intuitive: a smaller $f_{\rm split}$ implies a likely gap between two isolated peaks of the PDF (see, for instance, the top right panel of \cref{fig:interp}), which can be better fitted with two exponential tails extending from both sides.
Our PCHIP-ET scheme incorporates the inductive bias that for most SBI problems the tails of probabilistic distributions can be well modeled by Gaussians; if this is not the case, one may replace the Gaussian with e.g. student's $t$ or Cauchy for long-tailed distributions.

\section{Weights in $\mathcal{L}_0$}

\label{app:wxl0}

In this work, we use NQE to predict the quantiles equally spaced between $[0,\,1]$, which tends to put more emphasis on the regions with larger PDF where the neighboring quantiles are closer to each other, leading to potential instability in the tail regions.
Instead of directly weighting the different terms in $\mathcal{L}_0$, we adopt the following dropout strategy: for each training batch, we only keep $0<p_0\leq1$ of the terms in $\mathcal{L}_0$ using a no-replacement multinomial sampling with weights proportional to $\left<p_{\rm}\right>_{\rm avg}^{-f_0}$, $\,\left<p_{\rm}\right>_{\rm avg} \equiv (\left<p_{\rm}\right>_{\rm left} + \left<p_{\rm}\right>_{\rm right}) / 2$, with $p_0=0.5$ and $f_0=1$ by default.
This will effectively upweight the quantiles where the PDF is small, while the no-replacement sampling prevents specific terms from having too large weights that dominate the whole loss function.

\section{Benchmark Problems}

\label{app:prob}

We use the following problems from \citet{lueckmann2021benchmarking} to benchmark the performance of the SBI methods.
The ``ground truth" posterior samples are available for all the problems.

\subsection{Two Moons (TM)}

A toy problem with complicated global (bimodality) as well as local (crescent shape) structures.

{\setlength{\extrarowheight}{5pt}
\begin{tabularx}{\textwidth}{@{}ll@{}}
    \textbf{Prior} & %
        $\mathcal{U}(-{\bm 1}, {\bm 1})$ \\
    \textbf{Simulator} & %
        \bigcell{l}{
            $ \boldsymbol{x} | \btheta = \begin{bmatrix}
              r \cos(\alpha)+0.25\\
              r \sin(\alpha)
            \end{bmatrix} + $
            $\begin{bmatrix}
              {-|\theta_1+\theta_2|}/{\sqrt{2}} \\
              {(-\theta_{1}+\theta_{2}})/{\sqrt{2}} \\
            \end{bmatrix} $,\,
            where $ \alpha \sim \mathcal{U}(-\pi/{2}, \pi/2) $,
            $ r \sim \mathcal{N}(0.1,0.01^{2}) $
        } \\
    \textbf{Dimensionality} & %
        ${\btheta} \in \mathbb{R}^2, {\bx} \in \mathbb{R}^2$ \\
    \textbf{References} & %
        \citet{greenberg2019automatic}
\end{tabularx}}

\subsection{SLCP with Distractors (SLCPD)}

A challenging problem designed to have a simple likelihood and a complex posterior, with uninformative dimensions (distractors) added to the observation.

{\setlength{\extrarowheight}{5pt}
\begin{tabularx}{\textwidth}{@{}ll@{}}
    \textbf{Prior} & %
        $\mathcal{U}(-{\bm 3}, {\bm 3})$ \\
    \textbf{Simulator} & %
        \bigcell{l}{
            $ \bx|\btheta = (\bx_{1}, \ldots, \bx_{100})$,
            $\bx = p(\by)$, \\
            where $p$ re-orders the dimensions of $\by$ with a fixed random permutation; \\
            $ \by_{[1:8]} \sim \mathcal{N}({\bm m}_{\bm \theta}, {\bm S}_{\bm \theta}) $,
            $ \by_{[9:100]} \sim \frac{1}{20}\sum_{i=1}^{20} t_2({\bm \mu}^i,{\bm \Sigma}^i)$, \\
            where $ {\bm m}_{\btheta} = \begin{bmatrix} {\theta_{1}} \\ {\theta_{2}} \end{bmatrix} $,
            $ {\bm S}_{\btheta} = \begin{bmatrix} {s_{1}^{2}} & {\rho s_{1} s_{2}} \\ {\rho s_{1} s_{2}} & {s_{2}^{2}}\end{bmatrix} $,
            $ s_{1}=\theta_{3}^{2}, s_{2}=\theta_{4}^{2}, \rho=\tanh \theta_{5}$,
            $ {\bm \mu}^i \sim \mathcal{N}(0,15^2 {\bm I})$, \\
            $ {\bm \Sigma}_{j,k}^i \sim \mathcal{N}(0,9)$ for $j>k$,
            $ {\bm \Sigma}_{j,j}^i = 3 e^a$ with $a \sim \mathcal{N}(0,1)$,
            $ {\bm \Sigma}_{j,k}^i = 0$ otherwise
        } \\
    \textbf{Dimensionality} & %
        $\btheta \in \mathbb{R}^5, \bx \in \mathbb{R}^{100}$ \\
    \textbf{References} & %
        \citet{greenberg2019automatic}
\end{tabularx}}

\subsection{Bernoulli GLM Raw (BGLMR)}

Inference of a 10-parameter Generalized Linear Model (GLM) with raw Bernoulli observations.

{\setlength{\extrarowheight}{5pt}
\begin{tabularx}{\textwidth}{@{}ll@{}}
    \textbf{Prior} &
        \bigcell{l}{
            $\beta \sim \mathcal{N}(0,2)$,\,
            ${\bf f} \sim \mathcal{N}({\bm 0}, ({\bm F}^{\top} {\bm F})^{-1})$, \\
            ${\bm F}_{i,i-2} = 1$,\, ${\bm F}_{i,i-1} = -2$,\,
            ${\bm F}_{i,i} = 1+\sqrt{\frac{i-1}{9}}$,\,
            ${\bm F}_{i,j} = 0$ otherwise,\,
            $1\leq i,j \leq9$
        }\\
    \textbf{Simulator} &
        \bigcell{l}{
        ${\bx}|{\btheta} = ({\bx}_{1}, \ldots, {\bx}_{100})$,\,
        $x_i \sim \mathrm{Bern}(\eta(\mathbf{v}_i^{\top} {\bf f} + \beta))$,\,
        $\eta(\cdot)=\exp(\cdot)/(1 + \exp(\cdot))$ \\
        frozen input between time bins $i-8$ and $i$:
        ${\bm v}_i \sim \mathcal{N}({\bm 0},{\bm I})$
        } \\
    \textbf{Dimensionality} &
        ${\btheta} \in \mathbb{R}^{10}, {\bx} \in \mathbb{R}^{100}$ \\
    \textbf{Fixed Parameters} &
        Duration of task $T=100$\\
\end{tabularx}}

\subsection{Gaussian Mixture (GM)}

Inferring the common mean of a mixture of two Gaussians, one with much broader covariance than the other.

{\setlength{\extrarowheight}{5pt}
\begin{tabularx}{\textwidth}{@{}ll@{}}
    \textbf{Prior} & %
        $\mathcal{U}(-{\bm 1\bm 0}, {\bm 1\bm 0})$ \\
    \textbf{Simulator} & %
        \bigcell{l}{
            $ {\bx}|{\btheta} \sim 0.5 \; \mathcal{N}({\bx}|{\bm m}_{\btheta}={\btheta}, {\bm S} = {\bm I}) + $ $0.5 \; \mathcal{N}({\bx}|{\bm m}_{\btheta}={\btheta}, {\bm S} = 0.01 \odot {\bm I}) $
        } \\
    \textbf{Dimensionality} & %
        $\btheta \in \mathbb{R}^{2}, \bx \in \mathbb{R}^{2}$ \\
    \textbf{References} & %
        \citet{sisson2007sequential,beaumont2009adaptive,toni2009approximate,simola2021adaptive}
\end{tabularx}}

\subsection{SIR}

An epidemiological model describing the numbers of individuals in three possible states: susceptible $S$, infectious $I$, and recovered or deceased $R$.

{\setlength{\extrarowheight}{5pt}
\begin{tabularx}{\textwidth}{@{}ll@{}}
    \textbf{Prior} &
        \bigcell{l}{
            $\beta \sim \text{LogNormal}(\log(0.4), 0.5)$,\,
            $\gamma \sim \text{LogNormal}(\log(1/8), 0.2)$
        } \\
    \textbf{Simulator} &
        \bigcell{l}{
            $ {\bx}|{\btheta} = (x_{1}, \ldots, x_{10})$,\,
            $x_i \sim \mathcal{B}(1000,\frac{I}{N}) $,\, where $I$ is simulated from \\
            $\frac{dS}{dt} = -\beta \frac{SI}{N}$ \\
            $\frac{dI}{dt} = \beta \frac{SI}{N}-\gamma I$ \\
            $\frac{dR}{dt} = \gamma I$
        } \\
    \textbf{Dimensionality} &
        ${\btheta} \in \mathbb{R}^2, {\bx} \in \mathbb{R}^{10}$ \\
    \textbf{Fixed Parameters} &
        \bigcell{l}{
            Population size $N=1000000$ and duration of task $T=160$ \\
            Initial conditions:
            $(S(0),I(0),R(0)) = (N-1,1,0)$
        } \\
    \textbf{References} &
        \citet{kermack1927contribution}
\end{tabularx}}

\subsection{Lotka-Volterra (LV)}

An influential ecology model describing the dynamics of two interacting species.

{\setlength{\extrarowheight}{5pt}
\begin{tabularx}{\textwidth}{@{}ll@{}}
    \textbf{Prior} &
        \bigcell{l}{
            $\alpha \sim \text{LogNormal}(-0.125, 0.5)$,\,
            $\beta \sim \text{LogNormal}(-3, 0.5)$, \\
            $\gamma \sim \text{LogNormal}(-0.125, 0.5)$,\,
            $\delta \sim \text{LogNormal}(-3, 0.5)$
        } \\
    \textbf{Simulator} &
        \bigcell{l}{
            $ {\bx}|{\btheta} = (\bx_{1}, \ldots, \bx_{10})$,\,
            ${\bx}_{1,i} \sim \text{LogNormal}(\log(X),0.1)$,\,
            ${\bx}_{2,i} \sim \text{LogNormal}(\log(Y),0.1)$,\\
            $X$ and $Y$ are simulated from\\
            $\frac{dX}{dt} = \alpha X - \beta X Y$ \\
            $\frac{dY}{dt} =  -\gamma Y + \delta X Y$
        } \\
    \textbf{Dimensionality} &
        ${\btheta} \in \mathbb{R}^4, {\bx} \in \mathbb{R}^{20}$ \\
    \textbf{Fixed parameters} &
        \bigcell{l}{
            Duration of task $T=20$\\
            Initial conditions: $(X(0),Y(0))= (30,1)$
        }\\
    \textbf{References} & \citet{lotka1920analytical}
\end{tabularx}}

\section{Details of the Cosmology Application}

\label{app:cosmo}

We run $10^4$ dark-matter-only Particle Mesh (PM) simulations with $128^3$ particles in $256^3$ Mpc/h$^3$ boxes using the \texttt{pmwd} code \citep{li2022pmwd,li2022differentiable}, and generate two $128^2$ projected overdensity fields $\delta({\bm x}) \equiv \rho({\bm x}) / \Bar{\rho}$ from each simulation by dividing the box into two halves along the $z$ axis as the observation data.
We use 80\% simulations for training, 10\% for validation, and 10\% for test. We evaluate the calibration of NQE with the validation data, and plot \cref{fig:wl,fig:wl-c} with the test data.
The model parameters are $\Omega_m$, the total matter density today, and $\sigma_8$, the RMS matter fluctuation today in linear theory, with uniform priors $0.1\leq\Omega_m\leq0.5$ and $0.5\leq\sigma_8\leq1.1$.

As a proof-of-concept example, we substitute the expensive cosmological hydrodynamic simulations with a post-processing scale-independent bias \footnote{Here the \textit{bias} means any deviation of the actual observed field with respect to the dark-matter-only density field.} model over the density fields from the dark-matter-only simulations, i.e. $\delta({\bm x})\to b\,\delta({\bm x})$ with $b=1.02$ \footnote{But we still require that $\delta({\bm x})>0$.}.
In other words, we train NQE with $b=1$ simulations but requires the inference to be unbiased for $b=1.02$, which is achieved via the calibration of NQE.
A ResNet \citep{he2016deep} with 10 convolutional layers is utilized as the embedding network for a more efficient inference with the high dimensional data.

\clearpage

\section{Convergence Test of Coverage Evaluation}

\label{app:cover}

\begin{figure*}[ht]
\vskip 0.2in
\begin{center}
\centerline{\includegraphics[width=0.9\textwidth]{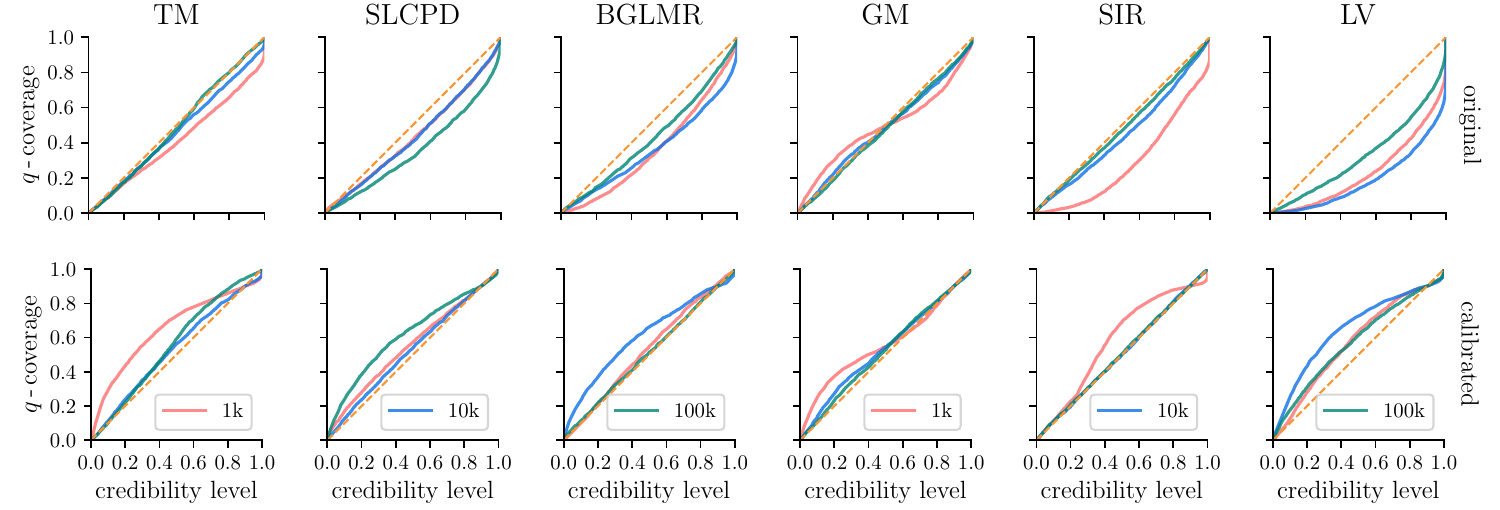}}
\caption{
Similar to \cref{fig:q-cover}, but using 2,000 simulations for the evaluation of $q-$coverage.\label{fig:q-cover-2k}}
\end{center}
\vskip -0.2in
\end{figure*}

\begin{figure*}[ht]
\vskip 0.2in
\begin{center}
\centerline{\includegraphics[width=0.9\textwidth]{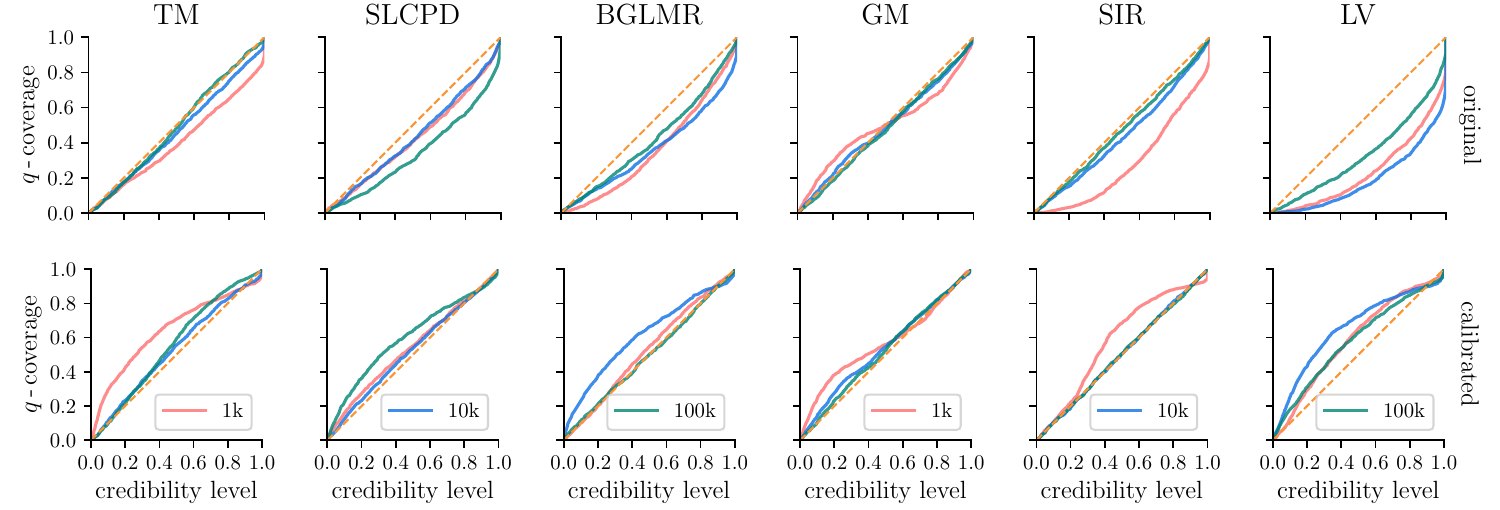}}
\caption{
Similar to \cref{fig:q-cover}, but using 1,000 simulations for the evaluation of $q-$coverage.\label{fig:q-cover-1k}}
\end{center}
\vskip -0.2in
\end{figure*}

\begin{figure*}[ht]
\vskip 0.2in
\begin{center}
\centerline{\includegraphics[width=0.9\textwidth]{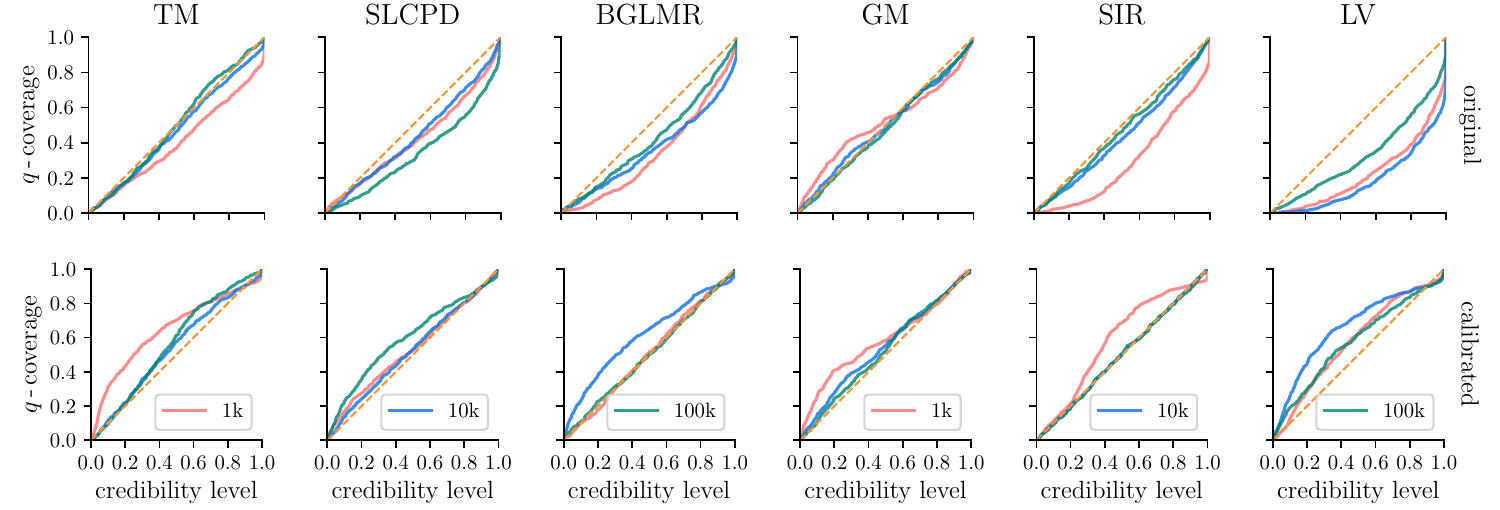}}
\caption{
Similar to \cref{fig:q-cover}, but using 500 simulations for the evaluation of $q-$coverage.\label{fig:q-cover-500}}
\end{center}
\vskip -0.2in
\end{figure*}

We check the convergence of the $q-$coverage evaluation in \cref{fig:q-cover-2k,fig:q-cover-1k,fig:q-cover-500}.
While \cref{fig:q-cover} in the main paper uses $10^4$ simulations to enhance the smoothness of the coverage curves, in most cases $\lesssim 10^3$ simulations should be sufficient for the evaluation of $q-$coverage.
Actually, the (pointwise) standard error of empirical coverage $p_{\rm ec}$ can be estimated using the properties of binomial distribution as $\Delta p_{\rm ec} = \sqrt{p_{\rm ec}\,(1-p_{\rm ec})\,/\,N_o}$, where $N_o$ is the number of simulations for the coverage evaluation \citep{sailynoja2022graphical}.
Therefore, with $N_o=10^3$, one has $\Delta p_{\rm ec}<1.6\%$ for all $p_{\rm ec}\in[0,\,1]$.

\clearpage

\section{Hyperparameter Choices}

\label{app:hyper}

\begin{table}[h]
\caption{Our baseline choice of NQE hyperparameters. \label{tab:hyper}} 
\begin{center}
\begin{tabular}{cc}
hyperparameter & value \\
\hline \vspace{-0.3cm} \\
$p_{\rm tl}$ & 0.6 \\
$p_0$ & $0.5$ \\
$f_0$ & $1.$ \\
$f_1$ & $1.1$ \\
$f_2$ & $0.8$ \\
$\lambda_{\rm reg}$ & 0.1 \\
\# of MLP hidden layers & 10 \\
\# of MLP hidden neurons per layer & 512 \\
$n_{\rm bin}$ & 16 \\
\end{tabular}
\end{center}
\end{table}

\begin{figure*}[ht]
\vskip 0.2in
\begin{center}
\centerline{\includegraphics[width=\textwidth]{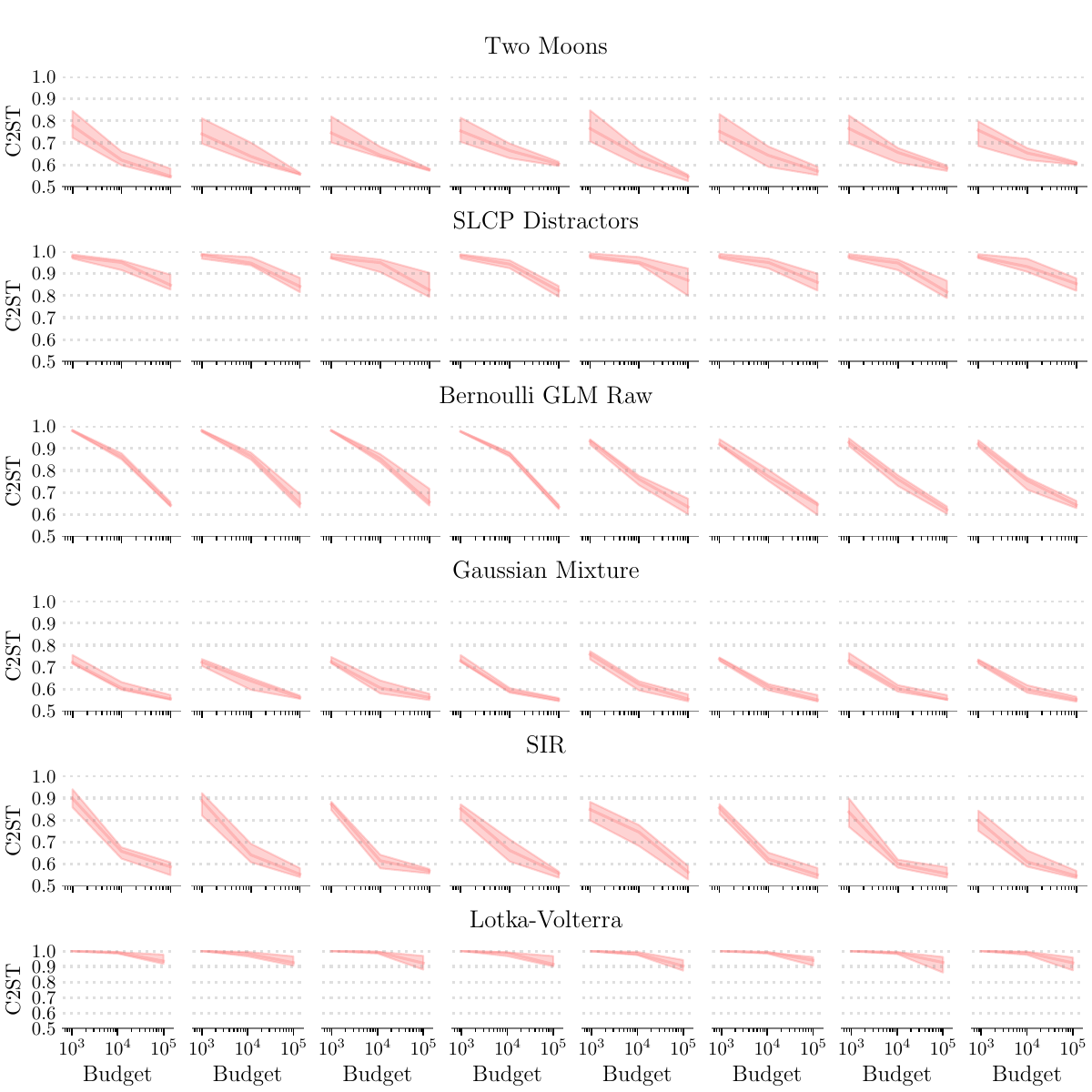}}
\caption{
A survey of NQE performance across different choices of hyperparameters.
From left to right, we set $f_0$ as (0, 0, 0, 0, 1, 1, 1, 1), and set $\lambda_{\rm reg}$ as (0, 0.01, 0.1, 1, 0, 0.01, 0.1, 1).
All other parameters are the same as \cref{tab:hyper}.
\label{fig:hyper0}}
\end{center}
\vskip -0.2in
\end{figure*}

\begin{figure*}[ht]
\vskip 0.2in
\begin{center}
\centerline{\includegraphics[width=\textwidth]{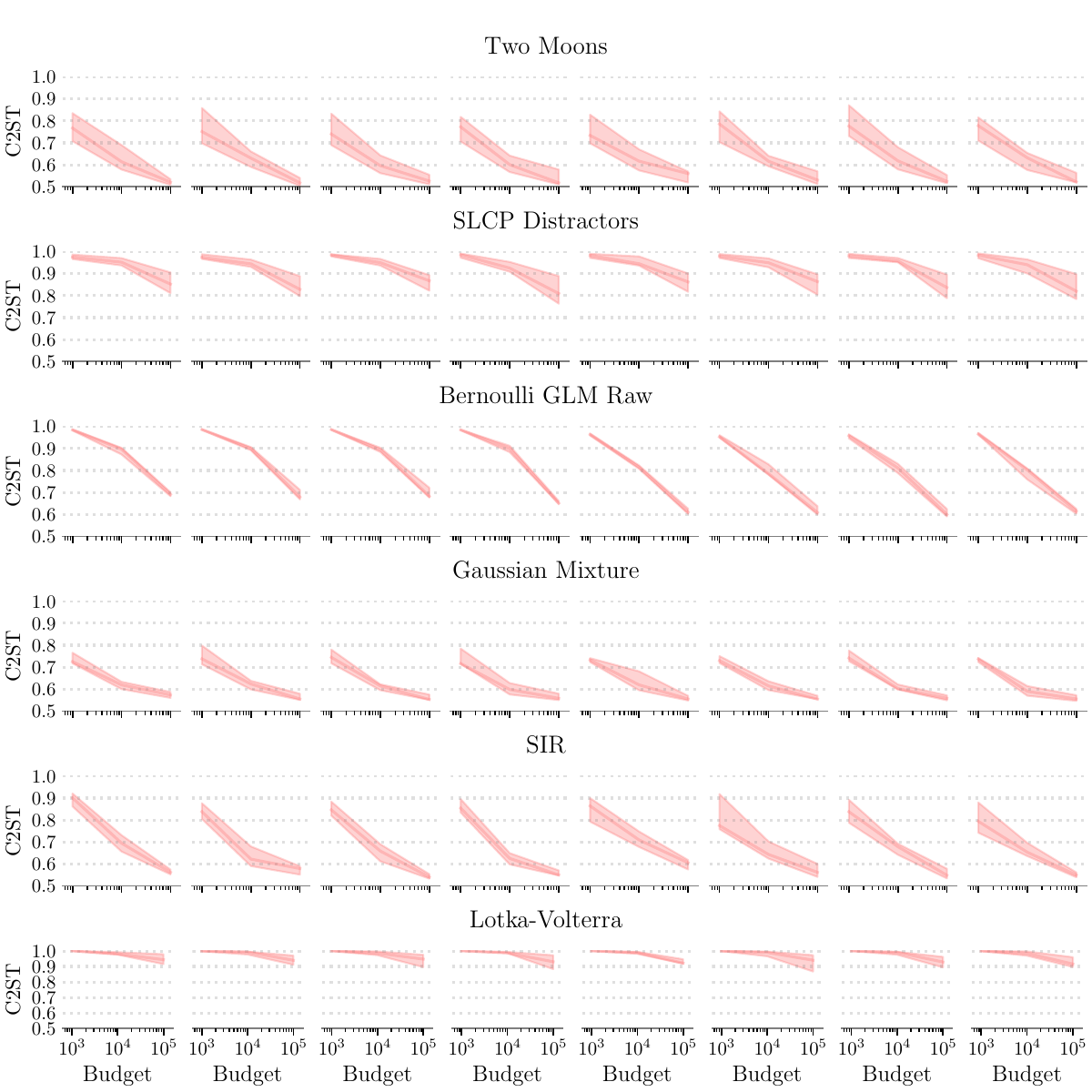}}
\caption{
Same as \cref{fig:hyper0}, but using 25 quantile bins. Increasing the number of bins is helpful for multimodal problems (e.g. TM) with large simulation budgets.
\label{fig:hyper1}}
\end{center}
\vskip -0.2in
\end{figure*}

We train all the models on NVIDIA A100 MIG GPUs using the AdamW optimizer \citep{loshchilov2017decoupled}, and find the wall time of NQE training to be comparable to existing methods like NPE.
Our PCHIP-ET scheme has been implemented with Cython \citep{behnel2010cython}, so that its evaluation is much faster than the quantile regression network calls for typical real-world examples.
We conduct a mild search for \{$f_0$, $\lambda_{\rm reg}$, $n_{\rm bin}$\} in \cref{fig:hyper0,fig:hyper1}, which leads to our baseline choice in \cref{tab:hyper}.
We reduce the stepsize by 10\% after every 5 epochs, and terminate the training if the loss does not improve after 30 epochs or when the training reaches 300 epochs.

We find that some tasks require a different stepsize while some tasks exhibit significant overfitting, so we train 9 realizations for each network with \{initial step size = 5e-4, 1e-4, 2e-5\} $\times$ \{AdamW weight decay = 0, 1, 10\}, and choose the realization with the smallest loss function.
Nevertheless, most problems have a clear preference regarding these two parameters so it should be straightforward to tune them for specific problems in practice.

\clearpage

\section{Additional Plots}

\label{app:plots}

\begin{figure*}[ht]
\vskip 0.2in
\begin{center}
\centerline{\includegraphics[width=0.95\textwidth]{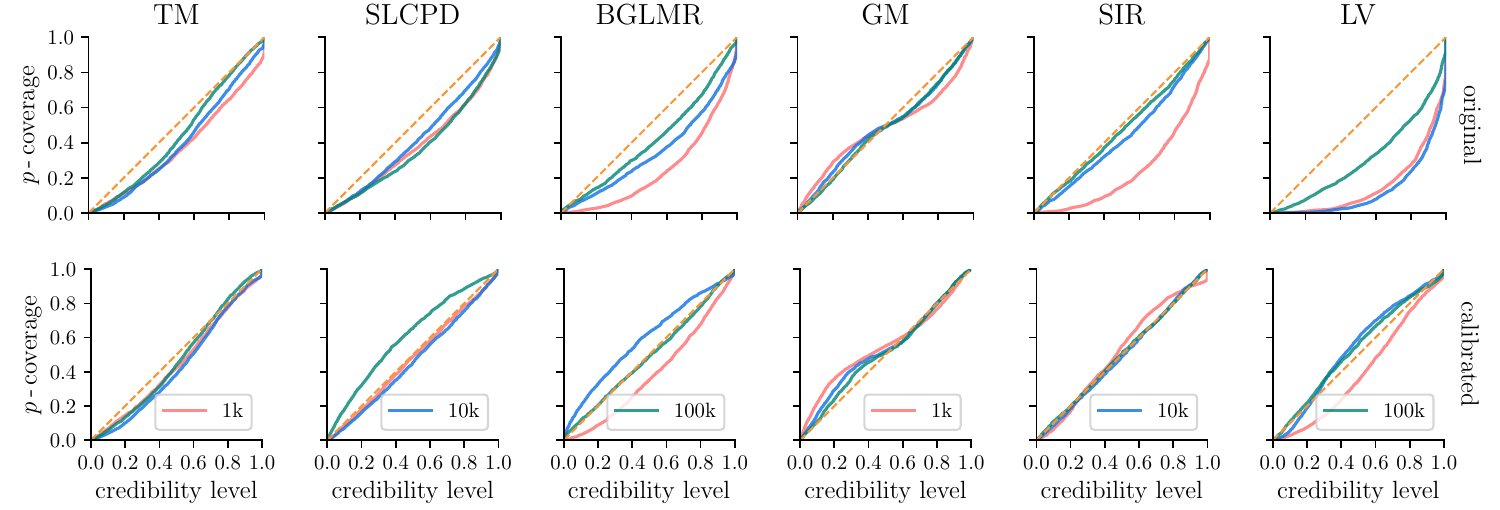}}
\caption{
Empirical coverage results using $p-$coverage, while the calibration is still evaluated using $q-$coverage. We find that the $p-$coverage results are qualitatively similar to the $q-$coverage in most cases, and the broadening calibration with $q-$coverage in the main text also mitigates the bias for the $p-$coverage. Nevertheless, one can always solve the broadening factor directly with $p-$coverage if one wishes the $p-$coverage to be strictly unbiased, at the cost of more network calls required than using $q-$coverage.
\label{fig:p-cover}}
\end{center}
\vskip -0.2in
\end{figure*}

\begin{figure*}[ht]
\vskip 0.2in
\begin{center}
\centerline{\includegraphics[width=0.85\textwidth]{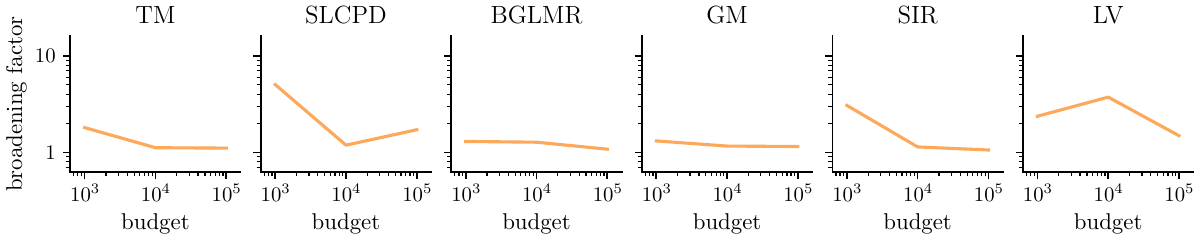}}
\caption{The actual broadening factor applied to remove the bias for the benchmark problems.
\label{fig:bf}}
\end{center}
\vskip -0.2in
\end{figure*}

\begin{figure}[t]
\vskip 0.2in
\begin{center}
\centerline{\includegraphics[width=0.5\textwidth]{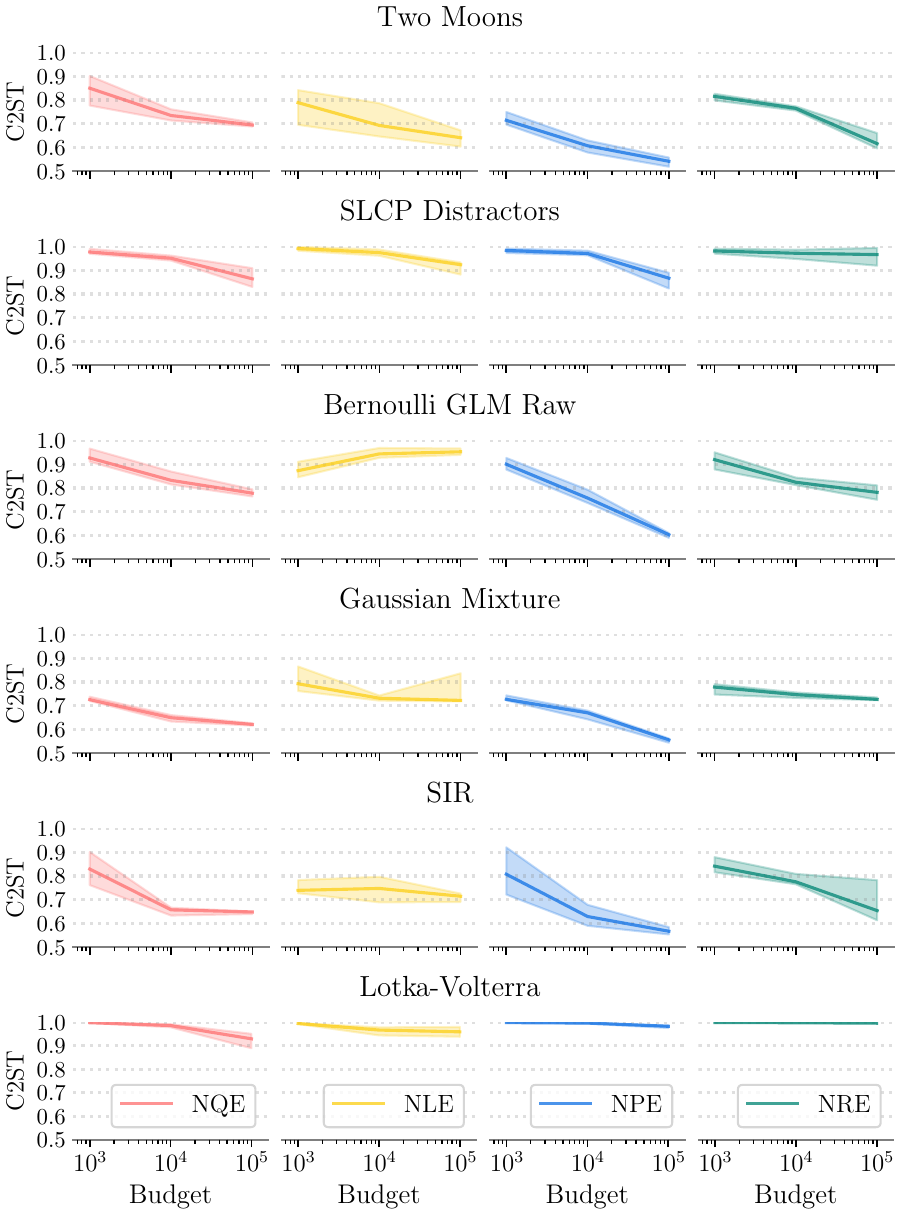}}
\caption{Similar to \cref{fig:02-ns}, but for NQE calibrated with the global broadening scheme. The C2ST of calibrated NQE is generally similar to or slightly worse than uncalibrated NQE in \cref{fig:02-ns}.\label{fig:c2st-b}}
\end{center}
\vskip -0.2in
\end{figure}

\vfill

\end{document}